\documentclass[runningheads]{llncs}

\usepackage{eccv}

\usepackage{eccvabbrv}

\usepackage{graphicx}
\usepackage{enumitem}
\usepackage{booktabs}
\usepackage{multirow}
\usepackage{array}
\usepackage{caption}
\usepackage{todonotes}
\usepackage{xcolor}

\usepackage[accsupp]{axessibility}  

\usepackage{subcaption}

\usepackage{hyperref}

\usepackage{orcidlink}

\usepackage{xspace}

\newcommand{\MultiPixmo}{Multi-PixMo\xspace}
\newcommand{\MultiPixmoCap}{Multi-PixMo-Cap\xspace}
\newcommand{\MultiPixmoAma}{Multi-PixMo-AskModelAnything\xspace}
\newcommand{\MultiPixmoCosyn}{Multi-PixMo-CoSyn-400k\xspace}

\newcommand{\Pixmo}{PixMo\xspace}
\newcommand{\PixmoCap}{PixMo-Cap\xspace}
\newcommand{\PixmoAma}{PixMo-AskModelAnything\xspace}
\newcommand{\PixmoCosyn}{CoSyn-400k\xspace}

\newcommand{\MultiPixmoCapShort}{\textsc{cap}\xspace}
\newcommand{\MultiPixmoAmaShort}{\textsc{ama}\xspace}
\newcommand{\MultiPixmoCosynShort}{\textsc{cos}\xspace}

\newcommand{\MultilingualBenchmarkSuiteLong}{Multimodal European Vision Benchmark}
\newcommand{\MultilingualBenchmarkSuite}{MEVBench\xspace}
\begin{document}

\title{Multilingual Training and Evaluation\\ Resources for Vision-Language Models}

\titlerunning{Multilingual Training and Evaluation Resources}



\author{Daniela Baiamonte\inst{1} \and Elena Fano\inst{1} \and Matteo Gabburo\inst{2} \and Stefano Simonazzi\inst{1} \and Leonardo Rigutini\inst{1} \and Andrea Zugarini\inst{1}}


\authorrunning{Baiamonte et al.}

\institute{Villanova.ai\inst{1}, Aithlas\inst{2} 
\email{\{dbaiamonte,efano,ssimonazzi,lrigutini,azugarini\}@villanova.ai, matteo.gabburo@aithlas.com} \\
\url{https://villanova.ai/}}



\maketitle

\begin{abstract}
 Vision Language Models (VLMs) achieved rapid progress in the recent years. However, despite their growth, VLMs development is heavily grounded on English, leading to two main limitations: (i) the lack of multilingual and multimodal datasets for training, and (ii) the scarcity of comprehensive evaluation benchmarks across languages. 
 In this work, we address these gaps by introducing a new comprehensive suite of resources for VLMs training and evaluation spanning five European languages (English, French, German, Italian, and Spanish). 
 We adopt a regeneration-translation paradigm that produces high-quality cross-lingual resources by combining curated synthetic generation and manual annotation.
 Specifically, we build \MultiPixmo, a training corpus obtained regenerating examples from \Pixmo pre-existing datasets with permissively licensed models: \PixmoCap, \PixmoAma, and \PixmoCosyn. On the evaluation side, we construct a set of multilingual benchmarks derived translating widely used English datasets (MMbench, ScienceQA, MME, POPE, AI2D). 
 We assess the quality of these resources through qualitative and quantitative human analyses, measuring inter-annotator agreement. Additionally, we perform ablation studies to demonstrate the impact of multilingual data, with respect to English only, in VLMs training. Experiments, comprising 3 different models, show that using multilingual, multimodal examples for training VLMs aids is consistently beneficial on non-English benchmarks, with positive transfer to English as well.

\end{abstract}

\section{Introduction}
\label{sec:intro}

Recent advances in Vision–Language Models (VLMs) have led to strong improvements across tasks such as image captioning, visual question answering, and multimodal reasoning~\cite{radford2021clip,li2022blip,alayrac2022flamingo}. Despite this rapid progress, the development of VLMs remains mainly in English. This imbalance introduces two critical limitations: (i) the scarcity of large-scale multilingual multimodal datasets for training, and (ii) the lack of comprehensive cross-lingual evaluation benchmarks.

Although multilingual text-only resources have become widely available~\cite{conneau2020xlmrlarge}, comparable multimodal datasets remain limited in scale, linguistic coverage, or annotation consistency~\cite{elliott2016multi30k}. As a consequence, multilingual VLMs are often trained primarily on English data and evaluated on English-centric benchmarks. 
%
Moreover, many widely adopted multimodal datasets rely on content generated using proprietary large language models and distributed under restrictive terms, making downstream reuse legally constrained~\cite{birhane2021datasets,paullada2021datacentric}.

In this work, we address both challenges by introducing a comprehensive multilingual suite for VLM training and evaluation spanning five European languages: English, French, German, Italian, and Spanish. Our approach combines controlled regeneration and translation strategies to produce legally compatible, high-quality cross-lingual multimodal resources.
On the training side, we present \MultiPixmo, a multilingual regeneration of three datasets from the \Pixmo suite\cite{deitke2024molmo}: \PixmoCap, \PixmoAma, and \PixmoCosyn\cite{yang2025scalingtextrichimageunderstanding}. Rather than directly translating the original annotations, we adopt a regeneration–translation paradigm in which original images and texts serve as semantic anchors, and a permissively licensed multimodal model is used to re-author captions, questions, and answers across languages. Importantly, English annotations are also regenerated under the same modeling assumptions, ensuring linguistic consistency and legal homogeneity across the entire corpus. The output of this process are three new datasets named \MultiPixmoCap, \MultiPixmoAma, and \MultiPixmoCosyn (for convenience abbreviated in \MultiPixmoCapShort, \MultiPixmoAmaShort and \MultiPixmoCosynShort).
On the evaluation side, we construct multilingual benchmarks by translating widely used English multimodal datasets (MM-bench, ScienceQA, MME, POPE, AI2D), enabling systematic cross-lingual assessment of VLM performance. Together, these resources provide a unified framework for multilingual training and evaluation under consistent licensing conditions.


Our contributions are threefold: (1) we introduce a large-scale multilingual suite for VLM training and evaluation covering five languages under unified, permissive licensing; (2) we propose a controlled multimodal regeneration framework that balances semantic fidelity, linguistic quality, and legal reusability and (3) we provide extensive human evaluation and empirical evidence showing that multilingual training data improves performance on non-English multimodal benchmarks.
By jointly addressing multilingual coverage, evaluation diversity, and licensing constraints, our work provides a practical and scalable foundation for open multilingual vision–language research.
\\
Our training and evaluation datasets are public and available~\footnote{https://huggingface.co/collections/VillanovaAI/multi-pixmo}\footnote{https://huggingface.co/collections/VillanovaAI/mevbench}.

\begin{figure}[t]
\centering
\includegraphics[width=\textwidth]{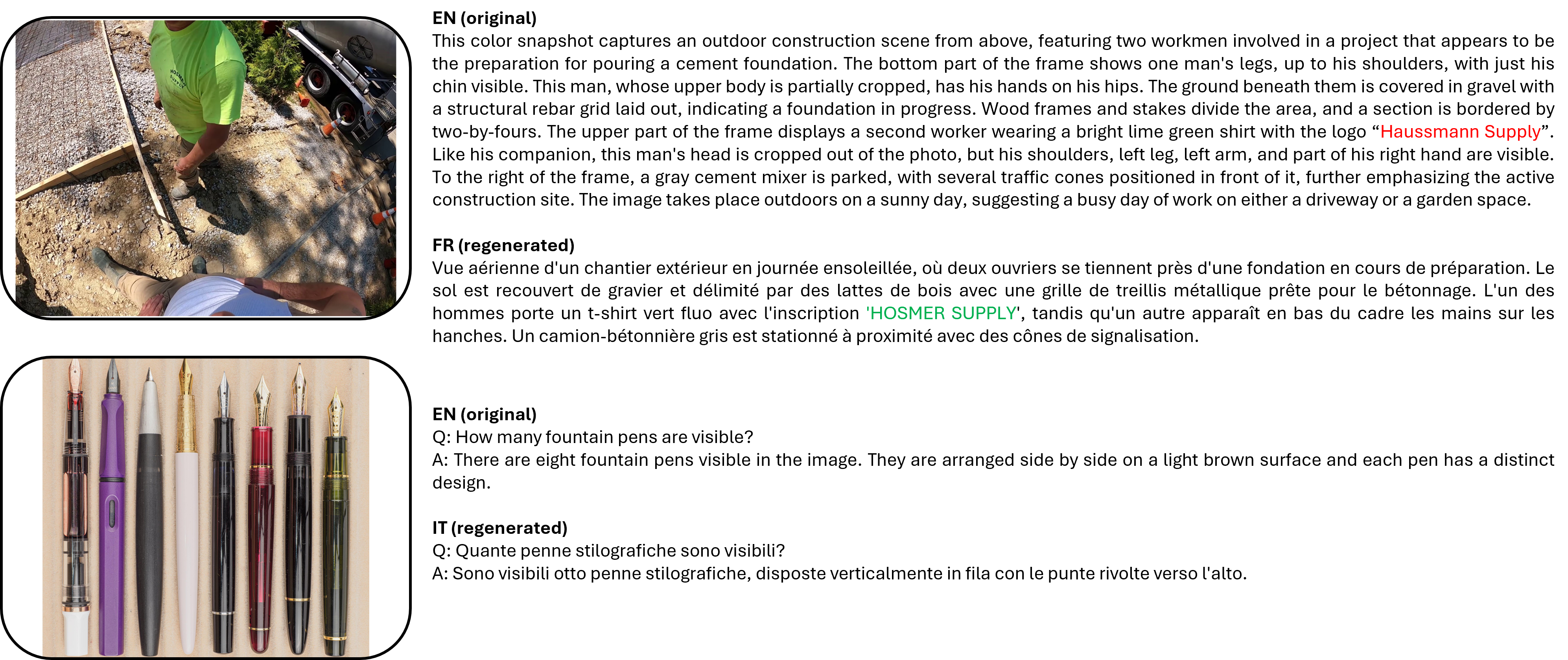}
\caption{Examples from \MultiPixmoCapShort (top) and \MultiPixmoAmaShort (bottom). The original \Pixmo annotation in English is shown first, followed by our regenerated version: French caption for \MultiPixmoCapShort; QA pair in Italian for \MultiPixmoAmaShort.}
\label{fig:qualitative_examples}
\end{figure}

\section{Related Works}
\label{sec:related_works}

\paragraph{\textbf{Training Corpora for Vision Language Models.}}
The development of modern Vision Language Models (VLMs) has been largely enabled by large-scale image-text and instruction-tuning corpora. Contrastive pretraining approaches such as CLIP~\cite{radford2021clip} demonstrated the effectiveness of web-scale supervision, leveraging noisy multilingual text-image pairs without curated question-answer alignment. 
Although few open models fully disclose their complete training pipelines, several public resources provide insight into current practices. LLaVA-OneVision~\cite{llavaonevisiondata} releases a large mixture including visual instruction data and academic benchmarks. InternVL builds upon curated multimodal mixtures such as MMPR-v1.2~\cite{mmprv12}. The IDEFICS family relies in part on OBELICS~\cite{laurenccon2023obelics}, a large-scale interleaved image-text corpus filtered from the web. SmolVLM and related open efforts (e.g., The Cauldron, Docmatix)~\cite{thecauldron,docmatix} aggregate instruction-tuning datasets spanning captioning, reasoning, chart understanding, and document analysis. Additional resources target specific tasks, such as rendered or OCR-rich text images~\cite{renderedtext}, as well as video-language corpora including FineVideo and LLaVA-Video-178K~\cite{finevideo,llavavideo}.
While these collections significantly diversify multimodal supervision, multilingual coverage is often low and unbalanced
and most instruction datasets remain in English~\cite{birhane2021datasheets,paullada2021datacentric}. 

In this context, we present \MultiPixmo, a multilingual VLM instruction tuning dataset for English, French, German, Italian, and Spanish. Built on a unified protocol rather than fragmented corpora, it focuses on controlled generative tasks to enable consistent cross-lingual alignment and robust open-ended reasoning.

\paragraph{\textbf{Multimodal Benchmarks and Multilingual Evaluation.}}
On the evaluation side, widely used multimodal benchmarks such as VQA v2~\cite{Goyal2017VQA2}, GQA~\cite{antol2015vqa}, MMBench~\cite{liu2023mmbench}, and AI2D~\cite{kembhavi2016ai2d} have become standard testbeds for visual reasoning and question answering. 
More recently, the scope of evaluation has expanded beyond generic VQA to encompass text-rich and document-centric scenarios. TextVQA~\cite{singh2019textvqa} requires reasoning over scene text, ChartQA~\cite{masry2022chartqa} focuses on numerical and logical reasoning over charts, and DocVQA~\cite{mathew2021docvqa} addresses structured document understanding. 
ScienceQA~\cite{Lu2022ScienceQA} evaluates multimodal scientific reasoning across school-level questions, while POPE~\cite{Li2023POPE} probes object hallucination by testing whether model outputs are visually grounded. MME~\cite{Fu2023MME} provides a comprehensive assessment of perception and cognition abilities through diverse multimodal tasks.
Complementary resources such as AI2D~\cite{kembhavi2016ai2d} and comprehensive suites like MMBench~\cite{liu2023mmbench} further stress diagram understanding, multi-step reasoning, OCR integration, and knowledge grounding, reflecting the increasing breadth of capabilities expected from modern VLMs.
Additionally, recent VLMs such as Qwen-VL~\cite{qwen3vl}, Gemma~\cite{gemma3} and SmolVLM~\cite{smolvlm} typically report performance on composite benchmark suites combining VQA-style reasoning, OCR-intensive tasks, chart interpretation, and document understanding.

However, most available datasets and benchmarks remain predominantly English-centric, with only a limited number of resources explicitly designed for multilingual evaluation. Although several benchmarks, such as MTVQA~\cite{tang2024mtvqa}, ALMBench~\cite{li2024almbench}, CVQA~\cite{liu2023cvqa}, and MMMB~\cite{zheng2024mmmb}, aim to assess multilingual visual understanding across languages, their coverage of European languages remains limited. 
To address this limitation, we introduce a \emph{multilingual benchmark suite} named \MultilingualBenchmarkSuite (\MultilingualBenchmarkSuiteLong) constructed by systematically translating and aligning a selection of widely adopted VLM evaluation datasets (MMBench, AI2D, ScienceQA, POPE, and MME) into five languages (English, French, German, Italian and Spanish).

\section{Training Resources}
\label{sec:multilingual_pixmo}

In this Section we present \MultiPixmoCap, \MultiPixmoAma, and \MultiPixmoCosyn, three multilingual datasets derived from \Pixmo~\cite{deitke2024molmo,yang2025scalingtextrichimageunderstanding} covering captioning, open-ended VQA, and text-rich image reasoning. 

We will also refer to them respectively as \MultiPixmoCapShort, \MultiPixmoAmaShort and \MultiPixmoCosynShort.
Rather than translating the original annotations, we adopt a regeneration/translation paradigm in which we leverage a permissively licensed multimodal model to rewrite all the examples of the original dataset, conditioning on both the image and the source text and ensuring linguistic quality and legal homogeneity. The resulting corpus spans over 1.1M images and 2.8M examples across five languages (Table~\ref{tab:language_distribution_combined}). 
In Sec.~\ref{ssec:regeneration_pipeline} we present our data generation pipeline, explaining in detail how we generate these new datasets, while in Sections \ref{ssec:quality_assessment}, \ref{ssec:qualitative_analysis} we provide quantitative and qualitative analysis of the data. The agreement between annotators is discussed in Section \ref{ssec:annotators_agreement}.

\subsection{Data Generation}
\label{ssec:regeneration_pipeline}




Our pipeline combines two strategies: translation, which converts text across languages without visual grounding, and regeneration, which conditions a multimodal model on both the image and the source text. We apply translation only to linguistically self-contained inputs (e.g. questions), and regeneration wherever visual grounding can improve quality, allowing the model to verify and disambiguate from the image. 
We perform all data generation using the permissively licensed \texttt{Qwen3-VL-235B-A22B-Instruct}~\cite{qwen3vl}. To ensure high-quality prompts, we involve native and expert speakers in a systematic refinement pipeline consisting of iterative generation cycles followed by blind reviews from independent QA annotators, until convergence to an optimal configuration.
We generated all three datasets with a language distribution of 40\% English, while French, German, Italian, and Spanish each account for 15\%.

\paragraph{\textbf{\MultiPixmoCap.}} Each image and its original English speech-to-text transcription are jointly provided to the model, which regenerates the caption in the target language under a strict grounding constraint: the image may only be used to verify or disambiguate content present in the transcription, not to introduce new information. This design choice allows to mitigate transcription noise while preserving the fidelity to the original human description. The same protocol is applied to English ensuring uniformity across languages. The resulting dataset contains 655,772 captions over 655,306 images (Table~\ref{tab:language_distribution_combined}).

\paragraph{\textbf{\MultiPixmoAma.}} The original human-authored questions are translated into each target language; answers are then regenerated by the multimodal model conditioned on the image and the translated question. English answers are also regenerated from the original questions, ensuring uniform modeling assumptions and licensing conditions across languages. During development, we found that ${\sim}$40\% of counting-related answers contained numerical errors; all these entries were manually corrected by annotators with access to the image. Counting-related entries are excluded from the quality evaluation (Section~\ref{ssec:quality_assessment}) to avoid confounding effects from this systematic error category. The final dataset comprises 124,714 QA triples (Table~\ref{tab:language_distribution_combined}).

\paragraph{\textbf{\MultiPixmoCosyn.}} \PixmoCosyn~\cite{yang2025scalingtextrichimageunderstanding} provides text-rich synthetic images (charts, diagrams, tables, etc.) whose original QA pairs were generated from code without access to the rendered image, making them unsuitable as semantic anchors. For this reason, we adopt a pure regeneration setting: the model is shown each image and prompted to generate QA pairs directly in the target language. We filtered out empty or excessively long outputs, as both correlate with low-quality images inherited from the original rendering pipeline. The final dataset contains 365,615 images and over 2M QA pairs  (Table~\ref{tab:language_distribution_combined}).

\begin{table}[t]
\centering
\scriptsize
\caption{Example of Multi-CoSyn-400K. The figure shows our regenerated question-answer pairs in German, along with their English translation.}
\label{tab:qualitative_examples_cosyn}
\setlength{\tabcolsep}{2pt}
\renewcommand{\arraystretch}{0.9}
\tiny\selectfont
\begin{tabular}{>{\centering\arraybackslash}m{0.35\columnwidth} m{0.30\columnwidth} m{0.28\columnwidth}}
\hline
\small
\textbf{Image} & \small\textbf{Regenerated Example} & \small\textbf{English translation} \\
\hline
\includegraphics[width=0.37\columnwidth]{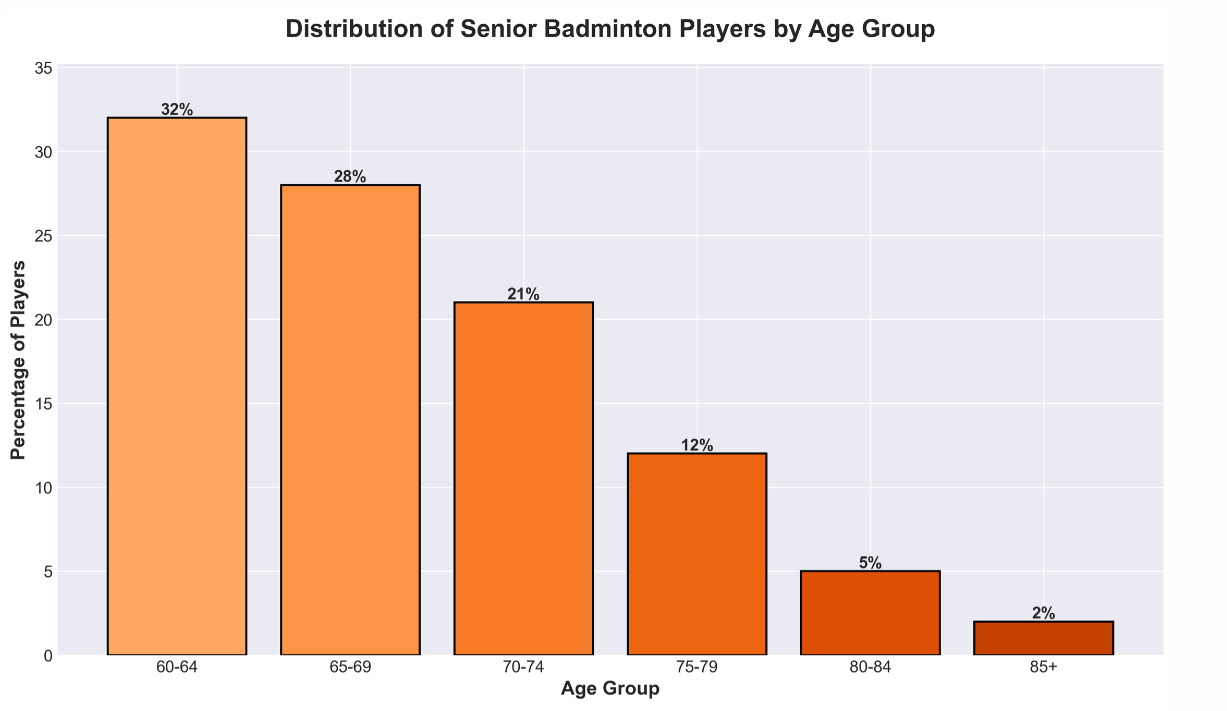}
&
{\tiny\selectfont\raggedright
\textbf{Q:} Was ist der Titel des Diagramms? \\
\textbf{A:} Distribution of Senior Badminton Players by Age Group \\
\textbf{Q:} Welche Altersgruppe hat den höchsten Prozentsatz an Spielern? \\
\textbf{A:} 60-64 \\
\textbf{Q:} Wie hoch ist der Prozentsatz der Spieler in der Altersgruppe 70-74? \\
\textbf{A:} 21\% \\
\textbf{Q:} Welche Altersgruppe hat den niedrigsten Prozentsatz an Spielern? \\
\textbf{A:} 85+ \\
\textbf{Q:} Wie viele Altersgruppen sind im Diagramm dargestellt? \\
\textbf{A:} 6 \\
}
&
{\tiny\selectfont\raggedright
\textbf{Q:} What is the title of the diagram? \\
\textbf{A:} Distribution of Senior Badminton Players by Age Group \\
\textbf{Q:} Which age group has the highest percentage of players? \\
\textbf{A:} 60-64 \\
\textbf{Q:} What is the percentage of players in the age group 70-74? \\
\textbf{A:} 21\% \\
\textbf{Q:} Which age group has the lowest percentage of players? \\
\textbf{A:} 85+ \\
\textbf{Q:} How many age groups are represented in the diagram? \\
\textbf{A:} 6 \\
}
\\
\hline
\end{tabular}
\end{table}

\begin{table}[h!]
\centering
\scriptsize
\caption{Language distribution across the three \MultiPixmo datasets.}
\label{tab:language_distribution_combined}
\begin{tabular}{l cc cc cc cc}
\toprule
\textbf{Lang} & \multicolumn{2}{c}{\textbf{\MultiPixmoCapShort}} & \multicolumn{2}{c}{\textbf{\MultiPixmoAmaShort}} & \multicolumn{2}{c}{\textbf{\MultiPixmoCosynShort}} & \multicolumn{2}{c}{\textbf{Total}} \\
\cmidrule(lr){2-3} \cmidrule(lr){4-5} \cmidrule(lr){6-7} \cmidrule(lr){8-9}
& Images & Examples & Images & Examples & Images & Examples & Images & Examples \\ 
\midrule
EN & 262,107 & 262,293 & 22,583 & 49,709 & 149,149 & 823,790 & 433,839 & 1,135,792 \\
IT & 98,287 & 98,354 & 8,482 & 18,774 & 54,742 & 302,372 & 161,511 & 419,500  \\
FR & 98,303 & 98,380 & 8,470 & 18,691 & 53,433 & 298,917 & 160,206 & 415,988  \\
ES & 98,313 & 98,376 & 8,520 & 18,877 & 56,073 & 314,255 & 162,906 & 431,508  \\
DE & 98,296 & 98,369 & 8,425 & 18,663 & 52,218 & 288,270 & 158,939 & 405,302  \\
\midrule
\textbf{Total} & \textbf{655,306} & \textbf{655,772} & \textbf{56,480} & \textbf{124,714} & \textbf{365,615} & \textbf{2,027,604} & \textbf{1,077,401} & \textbf{2,808,090} \\
\bottomrule
\end{tabular}
\end{table}

\subsection{Quality Assessment: Quantitative Analysis}
\label{ssec:quality_assessment}

To assess annotation quality, we conduct a human evaluation for each dataset and each language. The evaluation is carried out by two expert annotators per language.
Although the number of evaluators is limited, this setup prioritizes expertise and consistency, ensuring informed and reliable judgments.

\paragraph{\textbf{Annotation Guidelines.}} 
Each reviewer evaluated 125 examples, with all judgments being binary (correct/incorrect). The specific dimensions assessed vary by dataset.
For \textbf{\MultiPixmoCapShort}, annotators provided two judgments per example: (i) adherence of the caption to the image and (ii) linguistic correctness. For \textbf{\MultiPixmoAmaShort}, three judgments were needed: (i) translation quality of the question (accuracy and naturalness in the target language), (ii) content quality of the answer (factual correctness and consistency with image and question), and (iii) form quality of the answer. Separating content from form proved necessary, as annotations can be visually accurate but linguistically flawed, or vice versa, enabling finer-grained diagnosis of failure modes.
In \textbf{Multi-CoSyn-400K}, each image is associated with multiple question–answer pairs. However, to simplify the task for annotators, we ask them to evaluate each pair independently. The nature of the dataset, with short and concise questions and equally brief answers (often a single number or word), makes separate evaluation of content and form redundant, so the annotators were asked to give a single comprehensive quality judgment for each pair.\vspace{-0.2cm}

\paragraph{\textbf{Datasets Quality.}}
Results of the human evaluation are outlined in Table~\ref{tab:full_distribution_no_percent}. 
Overall, we can observe high quality in the linguistic form of the generated captions/answers, with percentages of correct evaluations beyond 96\% in most cases. Translating questions also leads to analogous results. Judgments about content adherence and faithfulness with respect to image present positive rating as well, usually above 85\%. Scores are slightly lower in French, but still in a positive range close to 80\%.
The quality of generated question and answer pairs in \MultiPixmoCosynShort is also above 90\% for all languages.

\begin{table} 
\centering
\footnotesize
\caption{Percentage of positive judgments per language and dimension (values in \%).}
\label{tab:full_distribution_no_percent}
\begin{tabular}{l cc ccc c}
\toprule
& \multicolumn{2}{c}{\textbf{\MultiPixmoCapShort}} & \multicolumn{3}{c}{\textbf{\MultiPixmoAmaShort}} & \multicolumn{1}{c}{\textbf{\MultiPixmoCosynShort}} \\
\cmidrule(lr){2-3} \cmidrule(lr){4-6} \cmidrule(lr){7-7}
Lang & Content & Form & Question & Content & Form & Content\\
EN & 83.2 & 99.6 & ---  & 82.4 & 99.2 & 93.1\\
IT & 96.4 & 99.6 & 95.6 & 87.6 & 91.2 & 95.6\\
ES & 96.8 & 100  & 96.8 & 89.2 & 96.4 & 93.2\\
FR & 79.2 & 96.4 & 97.6 & 77.2 & 96.4 & 93.7\\
DE & 92.0 & 95.2 & 99.6 & 92.8 & 98.8 & 91.2\\
\bottomrule
\end{tabular}
\end{table}

\subsection{Annotators Agreement}\label{ssec:annotators_agreement}
Among the 125 examples judged by each annotator, 50 were assigned to both annotators for each language to assess inter-annotator agreement.

\textbf{Metrics.} The inter-annotator agreement is measured with the Gwet’s AC1\cite{Gwet2008} metric. 
This choice is motivated by the nature of our task, which involves binary judgments and results in highly imbalanced label distributions.
Under these conditions, standard agreement metrics may yield unstable or misleading scores. 
The analysis shows consistently strong agreement among annotators (Table \ref{tab:agreement_all}) in all dataset and evaluation types, with scores typically varying from 0.8 to 1.00. Inspecting single languages, agreements in English, Italian and Spanish are particularly high. 
Disagreements are mainly concentrated in French, and to a lesser extent German, but still remaining in a good range of values.

\textbf{Results.} Qualitative inspection suggests that most of these differences tend to arise in borderline or ambiguity-sensitive cases, such as subjective visual interpretation like color descriptions (e.g., whether an object should be described as white or gray); minor spatial or perceptual ambiguities; differences in annotator tolerance for plausible but not fully verifiable details of the image.

\begin{table*} 
\centering
\footnotesize
\caption{Inter-annotator agreement (Gwet’s AC1) by dataset and annotation dimension: caption content and form for \MultiPixmoCapShort; question translation (Question), answer content (Content), and answer form (Form) for \MultiPixmoAmaShort; and QA pair correctness for \MultiPixmoCosynShort.}
\label{tab:agreement_all}

\begin{tabular}{l | ccc | cccc | c}
\toprule
& \multicolumn{3}{c|}{\textbf{\MultiPixmoCapShort}} 
& \multicolumn{4}{c|}{\textbf{\MultiPixmoAmaShort}} 
& \multicolumn{1}{c}{\textbf{\MultiPixmoCosynShort}} \\

Lang 
& \multicolumn{1}{c}{Content}
& \multicolumn{1}{c}{Form}
& \multicolumn{1}{c|}{Avg}

& \multicolumn{1}{c}{Question}
& \multicolumn{1}{c}{Content}
& \multicolumn{1}{c}{Form}
& \multicolumn{1}{c|}{Avg}

& \multicolumn{1}{c}{Content} \\

\midrule

EN 
& 0.81 & 1.00 & 0.91
& --- & 0.87 & 1.00 & 0.94
& 0.94 \\

IT 
& 0.98 & 0.98 & 0.98
& 0.93 & 0.90 & 0.88 & 0.90
& 0.95 \\

ES 
& 0.98 & 1.00 & 0.99
& 0.98 & 0.78 & 0.96 & 0.91
& 0.93 \\

FR 
& 0.68 & 0.89 & 0.79
& 0.98 & 0.73 & 0.89 & 0.87
& 0.93 \\

DE 
& 0.83 & 0.81 & 0.82
& 1.00 & 0.86 & 1.00 & 0.95
& 0.88 \\

\bottomrule
\end{tabular}
\end{table*}

\subsection{Quality Assessment: Qualitative Analysis}
\label{ssec:qualitative_analysis}

To complement the quantitative analysis, we performed a systematic inspection of negatively rated examples across all three datasets and languages, aiming to identify recurring error patterns and to trace each error to its source: the original data, known limitations of current VLMs, or our regeneration pipeline.



\paragraph{\textbf{Errors inherited from source data.}} A substantial portion of content errors does not originate from our pipeline but is inherited from upstream data. In \MultiPixmoCapShort, the input transcriptions are produced by an automatic speech-to-text system applied to spoken descriptions, which introduces noise, disfluencies, and occasional inaccuracies in references to visual content. Our regeneration model, constrained to ground its output in these transcriptions, propagates such artifacts when it cannot resolve them against the image. In \MultiPixmoCosynShort, image-level quality issues—such as cropped content, low-contrast labels, or excessively small fonts—stem from the original synthetic rendering pipeline~\cite{yang2025scalingtextrichimageunderstanding} and affect downstream QA generation. The qualitative analysis confirms this: image acceptability (about 88\%) sets some limits to QA correctness, as poorly rendered images limit the model's ability to produce accurate answers. It must be said that sometimes the model can generate meaningful QA pairs even when the image data is not clearly interpretable, mainly by focusing on surface details of the image.

\paragraph{\textbf{Errors reflecting known VLM limitations.}} The remaining content errors are largely consistent with well-documented limitations of current vision--language models~\cite{Li2023POPE,Fu2023MME}. Across all three datasets, we observe recurring issues with spatial reasoning (e.g., confusion between \textit{left} and \textit{right}), numerical counting, color identification under ambiguous lighting, and unverified inferences where the model fills gaps in the input with plausible but ungrounded details (Figure~\ref{fig:error_patterns}). These error categories are not specific to our regeneration setup; rather, they reflect the current capability frontier of multimodal models. Notably, for \MultiPixmoAmaShort, we identified counting-related questions as a systematic failure mode (approximately 40\% of counting answers contained numerical inaccuracies) and manually corrected all such entries, as described in Section~\ref{ssec:regeneration_pipeline}.

\paragraph{\textbf{Evidence of error correction.}} An important finding is that our multimodal regeneration setup can correct errors present in the original annotations. Since the model has access to both the transcription and the image, it can resolve transcription artifacts when visual evidence is unambiguous. Figure~\ref{fig:qualitative_examples} illustrates this: the text printed on a worker's shirt, inaccurately transcribed as ``Haussmann Supply'' in the original \MultiPixmoCapShort dataset, is correctly rendered as ``HOSMER SUPPLY'' in the regenerated French caption by leveraging the visual modality.

\paragraph{\textbf{Conciseness of regenerated annotations.}}
Besides specific error types, we observed that regenerated captions and answers tend to be shorter than the original ones. On English, the regenerated texts are on average 30–40\% shorter. This reflects the concise style of \texttt{Qwen3-VL-235B-A22B-Instruct}, which tends to omit secondary details. Nevertheless, the resulting captions remain substantially more descriptive than those found in traditional image-captioning datasets, often consisting of a single short sentence.

\begin{figure}[t]
\centering
\includegraphics[width=\textwidth]{images/4pics_final_2}
\caption{Unverified inferences in the French caption (\textit{are the crab's legs made of papier-mâché or not?}—first picture from the left) and in the Spanish answer (\textit{is chicken one of the ingredients?}—second picture from the left). Counting error (\textit{two feet} visible instead of one) in the French answer (third picture from the left) and ambiguous color (\textit{white or grey?}) in the German caption (first picture from the right).}
\label{fig:error_patterns}
\end{figure}

\section{Evaluation Benchmarks}
To assess the multilingual capabilities of our models, we constructed an evaluation suite named \MultilingualBenchmarkSuite (\MultilingualBenchmarkSuiteLong) made of benchmarks by extending widely adopted Vision--Language benchmarks. Additionally, we reserve held-out examples from \MultiPixmo. 
We selected a set of reference benchmarks that are commonly used in the literature and whose licenses allow redistribution and adaptation. The selected datasets cover different tasks and vision--language abilities. Concretely, we extended the following datasets: ScienceQA~\cite{Lu2022ScienceQA}, MMBench~\cite{liu2023mmbench}, AI2D~\cite{kembhavi2016ai2d}, POPE~\cite{Li2023POPE}, MME~\cite{Fu2023MME}.
Since these benchmarks are in English, we translated each example into French, German, Italian, and Spanish. Given the critical importance of evaluation quality, and the fact that these data are used exclusively for benchmarking already trained models, we prioritized translation accuracy. To this end, we employed state-of-the-art Large Language Models. In particular, we translated MMBench with \texttt{gemini-3-flash-preview}, AI2D with \texttt{gemini-2.5-flash}~\cite{comanici2025gemini} and MME, POPE and ScienceQA with \texttt{gpt-4.1-mini}~\cite{achiam2023gpt}.
The resulting multilingual benchmarks are referred to with the Multi- prefix. As summarized in Table~\ref{tab:benchmarks}, these datasets collectively comprise about 150k examples spanning multiple task formats, including multiple-choice questions (MCQ), counting tasks and binary (yes/no) classification. Exact Match (EM) is used as the primary metric for most tasks, while MME Score is adopted for Multi-MME in accordance with the original evaluation protocol.



\begin{table}[t]
\centering
\footnotesize
\caption{Multilingual Evaluation Benchmarks.}
\label{tab:benchmarks}
\begin{tabular}{llccc}
\toprule
 & \textbf{Benchmark} & \textbf{Task} & \textbf{Metric} & \textbf{\#Examples} \\
\midrule
\multirow{5}{*}{\textit{Translated}} 
& ScienceQA & MCQ & EM &  21,205 \\
& MMBench   & MCQ & EM & 21,645 \\
& AI2D      & MCQ & EM &  15,440\\
& POPE      & Y/N & EM &   45,000\\
& MME       & Y/N & MME Score &  11,870 \\
\bottomrule
\end{tabular}
\end{table}

\section{Experiments}
\label{sec:experiments}

We evaluate \MultiPixmo along different dimensions: (i) measuring whether multilingual training degrades English performance, (ii) quantifying the gains of multilingual training on non-English tasks. Additionally, we validate the generated annotations through supervised evaluation on held-out test splits of the datasets presented in Section~\ref{sec:multilingual_pixmo}.

\subsection{Experimental Setup}
\label{ssec:exp_setup}

We experiment with three language models, spanning on different sizes and architectures. In detail, we consider \texttt{LLaMA3-3B}~\cite{grattafiori2024llama}, \texttt{Qwen3-4B}~\cite{qwen3technicalreport}, and \texttt{LLaMA3-8B}~\cite{grattafiori2024llama}.

We trained the models following the LLaVA~\cite{liu2023llava} two-stage training recipe, and we conduct all the experiments isolating the contribution of the training data in order to better comprehend and analyze the benefits of each dataset in a more granular fashion. In detail, the LLaVA architecture is composed by three components which are the vision encoder, the projection layer and the language model. The first training stage (S1) trains a projection layer between the vision encoder and the language model usually using image-caption datasets, while the second stage (S2) trains the projection layer and the language model on tasks involving text and images (e.g., Vision Question Answering (VQA)). 

We use \MultiPixmoCapShort for S1, and \MultiPixmoAmaShort and \MultiPixmoCosynShort for S2, comparing the results against symmetrical baselines trained using the original English-only AllenAI datasets. In the supplementary materials we conduct an ablation on the relative importance of S1 and S2, and we confirm that S2 is the primary driver of multilingual gains, while multilingual caption alignment in S1 provides a consistent, but smaller, benefit. 

To isolate contribution of each resource, we define four training configurations, each specified as an S1$\,\to\,$S2 pipeline.

In particular, we denote each training run as an S1$\,\to\,$S2 pipeline, where S1 specifies the captioning dataset used for the projection layer alignment, and S2 the instruction-tuning mixture. We use \textsc{cap}, \textsc{ama}, and \textsc{cos} as shorthands for \MultiPixmoCap, \MultiPixmoAma, and \MultiPixmoCosyn, respectively, with subscripts $_{en}$ and $_{mu}$ indicating whether the English-only or the full multilingual version is used. For example, \textsc{cap}$_{mu}$$\,\to\,$\textsc{ama}$_{en}$ denotes multilingual alignment followed by English-only instruction tuning. 
In our experiments, we compare four settings: a monolingual baseline (\textsc{cap}$_{en}$$\,\to\,$\textsc{ama}$_{en}$), its multilingual counterpart (\textsc{cap}$_{mu}$$\,\to\,$\textsc{ama}$_{mu}$), and two extended configurations that augment S2 with CoSyn data (\textsc{cap}$_{en}$$\,\to\,$\textsc{ama}$_{en}$\,+\,\textsc{cos}$_{en}$ and \textsc{cap}$_{mu}$$\,\to\,$\textsc{ama}$_{mu}$\,+\,\textsc{cos}$_{mu}$). 
When combining \MultiPixmoAma with \MultiPixmoCosyn, we randomly subsample \MultiPixmoCosyn
to obtain a balanced 50/50 mixture.

\paragraph{\textbf{Training details.}} All models use the LLaVA architecture with SigLIP\footnote{\texttt{google/siglip-so400m-patch14-384}} as the vision encoder (384$\times$384 resolution, 729 image tokens) and a randomly initialized two-layer MLP projector. For S1, we use a constant learning rate of $1e-3$, an effective batch size of 256, and a maximum sequence length of 2048 tokens for 2 epochs. For S2, we use a cosine schedule with 200 warmup steps, a peak learning rate of $2e-5$, weight decay of $0.01$ and an effective batch size of 128. All training is conduceted in $bf16$ mixed precision on 8$\times$ NVIDIA H100 80GB GPUs. We perform the evaluation with VLMEvalKit \cite{duan2024vlmevalkit} using heuristic answer extraction (no LLM-as-a-judge) and greedy decoding. For AI2D, ScienceQA and MMBench, we report the accuracy, while for POPE we report F1 score on binary yes/no answers, while for MME, we used the original composite score.



\subsection{Results}

\paragraph{\textbf{Impact of Multilingual Training on English Benchmarks.}}

A natural concern when introducing multilingual data is whether it comes at the cost of English performance. Table~\ref{tab:results_english} addresses this question by reporting the performance on the English split of our translated benchmarks, measuring whether multilingual training causes regression on monolingual tasks. 

The results show that multilingual training does not degrade English accuracy. For example, considering only the \textsc{CAP}$\to$\textsc{AMA} settings, \texttt{LLaMA3-3B} differs by at most 0.5 points on almost every benchmark with the only exception of MME (where this difference is not noticeable due to the MME's composite score). \texttt{Qwen3-4B}, exhibits a similar pattern with the multilingual model matching the English one on AI2D and MMBench, even marginally improving on ScienceQA. Major improvements emerge for \texttt{LLaMA3-8B} where multilingual training yields gains of +5.4 on AI2D, +3.5 on ScienceQA, and +365.9 on MME. 
The same behavior holds when introducing \MultiPixmoCosynShort to the second stage of the training. \MultiPixmoCosynShort is not always beneficial for the performance on the benchmarks, but its relative comparison with English-only CoSyn-400K does introduce degradations.

\begin{table}
\centering
\scriptsize
\caption{English performance on our translated benchmarks under four training conditions: English-only ($en$), Multilingual ($mu$), and their enhanced versions with Multi-CoSyn-400K (+\textsc{cos}$_{mu}$). Best result per model group is \textbf{bold}.}
\label{tab:results_english}
\begin{tabular}{l l @{\;} c @{\;} l @{\;} c @{\;} lccccc}
\toprule
\textbf{Model} & \multicolumn{5}{c}{\textbf{Train}} & \textbf{AI2D} & \textbf{ScienceQA} & \textbf{MMBench} & \textbf{POPE} & \textbf{MME}\\
\midrule

\multirow{4}{*}{\texttt{LLaMA3-3B}}
 & \textsc{cap}$_{en}$ & $\to$ & \textsc{ama}$_{en}$ & &                         & 51.7 & 70.8 & 72.0 & 81.8 & 1213.4 \\
 & \textsc{cap}$_{mu}$ & $\to$ & \textsc{ama}$_{mu}$ & &                         & 51.2 & \textbf{70.9} & \textbf{72.1} & 82.3 & \textbf{1421.1} \\
 & \textsc{cap}$_{en}$ & $\to$ & \textsc{ama}$_{en}$ & $+$ & \textsc{cos}$_{en}$ & \textbf{53.0} & 70.0 & 71.0 & \textbf{84.8} & 1350.5 \\
 & \textsc{cap}$_{mu}$ & $\to$ & \textsc{ama}$_{mu}$ & $+$ & \textsc{cos}$_{mu}$ & 52.6 & 69.6 & 71.0 & \textbf{84.8} & 1350.5 \\
\midrule

\multirow{4}{*}{Qwen3-4B}
 & \textsc{cap}$_{en}$ & $\to$ & \textsc{ama}$_{en}$ & &                          & 64.3 & 77.1 & 80.2 & \textbf{85.4} & 1557.7 \\
 & \textsc{cap}$_{mu}$ & $\to$ & \textsc{ama}$_{mu}$ & &                          & 64.1 & 77.4 & 79.8 & 82.8 & 1498.4 \\
 & \textsc{cap}$_{en}$ & $\to$ & \textsc{ama}$_{en}$ & $+$ & \textsc{cos}$_{en}$  & 64.0 & \textbf{78.0} & \textbf{81.0} & 85.1 & \textbf{1681.7} \\
 & \textsc{cap}$_{mu}$ & $\to$ & \textsc{ama}$_{mu}$ & $+$ & \textsc{cos}$_{mu}$  & \textbf{64.4} & 77.6 & 80.6 & 85.1 & \textbf{1681.7} \\
\midrule

\multirow{4}{*}{LLaMA3-8B}
 & \textsc{cap}$_{en}$ & $\to$ & \textsc{ama}$_{en}$ & &                         & 52.1 & 68.5 & 75.1 & 81.4 & 1208.9 \\
 & \textsc{cap}$_{mu}$ & $\to$ & \textsc{ama}$_{mu}$ & &                         & \textbf{57.5} & \textbf{72.0} & \textbf{75.4} & \textbf{82.9} & \textbf{1574.8} \\
 & \textsc{cap}$_{en}$ & $\to$ & \textsc{ama}$_{en}$ & $+$ & \textsc{cos}$_{en}$ & 51.4 & 70.5 & 72.0 & 80.8 & 1257.1 \\
 & \textsc{cap}$_{mu}$ & $\to$ & \textsc{ama}$_{mu}$ & $+$ & \textsc{cos}$_{mu}$ & 55.9 & 68.3 & 73.9 & 81.5 & 1547.9 \\
\bottomrule
\end{tabular}%
\end{table}

\begin{table}
\centering
\scriptsize
\caption{Non-English performance on MCQ benchmarks. Models are evaluated under four training conditions: English-only ($en$), Multilingual ($mu$), and their versions with Multi-CoSyn-400K (+\textsc{cos}$_{mu}$). Results are averaged over the four languages. Best per model group is \textbf{bold}.}
\label{tab:results_multilingual_mcq}
\begin{tabular}{l l l @{\;} c @{\;} l @{\;} c @{\;} l ccccc}
\toprule
\textbf{Bench} & \textbf{Model} & \multicolumn{5}{c}{\textbf{Train}} & \textbf{DE} & \textbf{ES} & \textbf{FR} & \textbf{IT} & \textbf{Avg} \\
\midrule
\multirow{12}{*}{AI2D} 
& \multirow{4}{*}{LLaMA3-3B} 
& \textsc{cap}$_{en}$ & $\to$ & \textsc{ama}$_{en}$ & &                           & 42.4 & 44.4 & 45.2 & 44.0 & 44.0 \\
& & \textsc{cap}$_{mu}$ & $\to$ & \textsc{ama}$_{mu}$ & &                         & 44.3 & 46.3 & 47.7 & 45.6 & 46.0 \\
& & \textsc{cap}$_{en}$ & $\to$ & \textsc{ama}$_{en}$ & $+$ & \textsc{cos}$_{en}$ & 45.8 & 46.5 & 46.6 & 44.9 & 46.0 \\
& & \textsc{cap}$_{mu}$ & $\to$ & \textsc{ama}$_{mu}$ & $+$ & \textsc{cos}$_{mu}$ & \textbf{46.9} & \textbf{48.6} & \textbf{49.0} & \textbf{47.0} & \textbf{47.9} \\
\cmidrule{2-12}
& \multirow{4}{*}{Qwen3-4B} 
& \textsc{cap}$_{en}$ & $\to$ & \textsc{ama}$_{en}$ & &                           & 59.1 & 62.3 & 60.2 & 59.9 & 60.4 \\
& & \textsc{cap}$_{mu}$ & $\to$ & \textsc{ama}$_{mu}$ & &                         & 58.8 & 61.4 & 60.6 & 60.9 & 60.4 \\
& & \textsc{cap}$_{en}$ & $\to$ & \textsc{ama}$_{en}$ & $+$ & \textsc{cos}$_{en}$ & 59.8 & 62.7 & 61.7 & 61.1 & 61.3 \\
& & \textsc{cap}$_{mu}$ & $\to$ & \textsc{ama}$_{mu}$ & $+$ & \textsc{cos}$_{mu}$ & \textbf{60.3} & \textbf{63.4} & \textbf{62.0} & \textbf{63.4} & \textbf{62.3} \\
\cmidrule{2-12}
& \multirow{4}{*}{LLaMA3-8B} 
& \textsc{cap}$_{en}$ & $\to$ & \textsc{ama}$_{en}$ & &                 & 44.6 & 46.3 & 46.1 & 45.2 & 45.5 \\
& & \textsc{cap}$_{mu}$ & $\to$ & \textsc{ama}$_{mu}$ & &               & \textbf{49.9} & \textbf{52.6} & \textbf{52.6} & \textbf{51.9} & 51.8 \\
& & \textsc{cap}$_{en}$ & $\to$ & \textsc{ama}$_{en}$ & + & \textsc{cos}$_{en}$ & 45.0 & 47.7 & 47.8 & 46.1 & 47.6 \\
& & \textsc{cap}$_{mu}$ & $\to$ & \textsc{ama}$_{mu}$ & + & \textsc{cos}$_{mu}$ & 48.7 & 53.3 & 52.5 & 51.1 & \textbf{52.3} \\
\midrule
\multirow{12}{*}{ScienceQA} 
& \multirow{4}{*}{LLaMA3-3B} 
& \textsc{cap}$_{en}$ & $\to$ & \textsc{ama}$_{en}$ & &                           & 67.4 & \textbf{67.1} & 66.6 & 66.4 & 66.9 \\
& & \textsc{cap}$_{mu}$ & $\to$ & \textsc{ama}$_{mu}$ & &                         & \textbf{68.3} & 66.7 & \textbf{66.9} & \textbf{66.8} & \textbf{67.2} \\
& & \textsc{cap}$_{en}$ & $\to$ & \textsc{ama}$_{en}$ & $+$ & \textsc{cos}$_{en}$ & 67.3 & 66.2 & 65.6 & 66.2 & 66.3 \\
& & \textsc{cap}$_{mu}$ & $\to$ & \textsc{ama}$_{mu}$ & $+$ & \textsc{cos}$_{mu}$ & 65.4 & 65.5 & 64.7 & 64.8 & 65.1 \\
\cmidrule{2-12}
& \multirow{4}{*}{Qwen3-4B} 
& \textsc{cap}$_{en}$ & $\to$ & \textsc{ama}$_{en}$ & &                           & 74.4 & 74.3 & 75.9 & 75.0 & 74.9 \\
& & \textsc{cap}$_{mu}$ & $\to$ & \textsc{ama}$_{mu}$ & &                         & 75.1 & 73.7 & 76.3 & 75.6 & 75.2 \\
& & \textsc{cap}$_{en}$ & $\to$ & \textsc{ama}$_{en}$ & $+$ & \textsc{cos}$_{en}$ & 73.2 & 74.0 & 75.2 & 73.8 & 74.0 \\
& & \textsc{cap}$_{mu}$ & $\to$ & \textsc{ama}$_{mu}$ & $+$ & \textsc{cos}$_{mu}$ & \textbf{76.0} & \textbf{75.5} & \textbf{76.8} & \textbf{76.5} & \textbf{76.2} \\
\cmidrule{2-12}
& \multirow{4}{*}{LLaMA3-8B} 
& \textsc{cap}$_{en}$ & $\to$ & \textsc{ama}$_{en}$ & &                   & 65.6 & 65.8 & 65.2 & 65.0 & 65.4 \\
& & \textsc{cap}$_{mu}$ & $\to$ & \textsc{ama}$_{mu}$ & &                 & \textbf{68.3} & \textbf{67.8} & \textbf{68.9} & 66.6 & \textbf{67.9} \\
& & \textsc{cap}$_{en}$ & $\to$ & \textsc{ama}$_{en}$ & + & \textsc{cos}$_{en}$   & 66.9 & 67.2 & 66.3 & \textbf{67.4} & 67.7 \\
& & \textsc{cap}$_{mu}$ & $\to$ & \textsc{ama}$_{mu}$ & + & \textsc{cos}$_{mu}$   & 63.2 & 65.3 & 65.6 & 65.0 & 65.5 \\
\midrule
\multirow{12}{*}{MMBench} 
& \multirow{4}{*}{LLaMA3-3B} 
& \textsc{cap}$_{en}$ & $\to$ & \textsc{ama}$_{en}$ & &                           & 67.8 & 68.9 & 69.6 & 67.6 & 68.5 \\
& & \textsc{cap}$_{mu}$ & $\to$ & \textsc{ama}$_{mu}$ & &                         & \textbf{70.0} & \textbf{69.9} & \textbf{70.2} & \textbf{69.7} & \textbf{70.0} \\
& & \textsc{cap}$_{en}$ & $\to$ & \textsc{ama}$_{en}$ & $+$ & \textsc{cos}$_{en}$ & 67.6 & 67.7 & 68.1 & 66.5 & 67.5 \\
& & \textsc{cap}$_{mu}$ & $\to$ & \textsc{ama}$_{mu}$ & $+$ & \textsc{cos}$_{mu}$ & 68.7 & 69.0 & 69.5 & 68.6 & 69.0 \\
\cmidrule{2-12}
& \multirow{4}{*}{Qwen3-4B} 
& \textsc{cap}$_{en}$ & $\to$ & \textsc{ama}$_{en}$ & & & 78.4 & 78.0 & 77.9 & 79.0 & 78.3 \\
& & \textsc{cap}$_{mu}$ & $\to$ & \textsc{ama}$_{mu}$ & & & 78.8 & 79.0 & 79.0 & 79.2 & 79.0 \\
& & \textsc{cap}$_{en}$ & $\to$ & \textsc{ama}$_{en}$ & $+$ & \textsc{cos}$_{en}$ & 77.5 & 77.5 & 78.2 & 78.1 & 77.8 \\
& & \textsc{cap}$_{mu}$ & $\to$ & \textsc{ama}$_{mu}$ & $+$ & \textsc{cos}$_{mu}$ & \textbf{79.4} & \textbf{79.2} & \textbf{79.8} & \textbf{79.9} & \textbf{79.6} \\
\cmidrule{2-12}
& \multirow{4}{*}{LLaMA3-8B} 
& \textsc{cap}$_{en}$ & $\to$ & \textsc{ama}$_{en}$ & &                 & 72.0 & 72.9 & 72.6 & 71.8 & 72.3 \\
& & \textsc{cap}$_{mu}$ & $\to$ & \textsc{ama}$_{mu}$ & &               & \textbf{73.1} & \textbf{73.4} & \textbf{73.4} & \textbf{73.2} & \textbf{73.3} \\
& & \textsc{cap}$_{en}$ & $\to$ & \textsc{ama}$_{en}$ & + & \textsc{cos}$_{en}$ & 69.4 & 69.2 & 69.1 & 69.1 & 69.8 \\
& & \textsc{cap}$_{mu}$ & $\to$ & \textsc{ama}$_{mu}$ & + & \textsc{cos}$_{mu}$ & 72.8 & 72.7 & 72.2 & 72.1 & 72.7 \\
\bottomrule
\end{tabular}
\end{table}

\paragraph{\textbf{Impact of Multilingual Training on Multilingual Benchmarks.}}

Tables \ref{tab:results_multilingual_mcq} and \ref{tab:results_multilingual_yn} report per-language results on multiple languages (DE, ES, FR, IT), respectively, for Multiple Choice and Yes/No benchmarks tasks. 
Multilingual training show consistent improvements across all languages and models. For example, in Table~\ref{tab:results_multilingual_mcq}, MMBench \textsc{cap}$_{mu}$$\,\to\,$\textsc{ama}$_{mu}$ outperforms
\textsc{cap}$_{en}$$\,\to\,$\textsc{ama}$_{en}$ by +1.5~points on
average for \texttt{LLaMA3-3B}, +0.7 for \texttt{Qwen3-4B}, and +1.0 for \texttt{LLaMA3-8B}.
On AI2D, the gains range from +2.0 for \texttt{LLaMA3-3B} to
+6.3 for \texttt{LLaMA3-8B}, while \texttt{Qwen3-4B} shows equivalent
average performance (60.4) with minor per-language variations. Table \ref{tab:results_multilingual_yn} presents similar patterns for POPE and MME as well, with the exception of MME with \texttt{LLaMA3-3B} where English-only training outperforms the other strategies.
Including \MultiPixmoCosynShort to the training mixture yields additional gains on benchmarks aligned with the \MultiPixmoCosynShort domain. For example, we achieved the best scores on AI2D for both \texttt{LLaMA3-3B} and \texttt{Qwen3-4B}. This results is expected since \MultiPixmoCosynShort  is designed for diagrams/charts understanding. In contrast, \MultiPixmoCosynShort does not help on ScienceQA, \texttt{LLaMA3-3B} with
\textsc{cap}$_{mu}$$\,\to\,$\textsc{ama}$_{mu}$\,+\,\textsc{cos}$_{mu}$
drops by $-$2.1 points relative to
\textsc{cap}$_{mu}$$\,\to\,$\textsc{ama}$_{mu}$. This suggests that the
domain mismatch between synthetic chart data and science questions can be
detrimental for smaller models. When comparing the monolingual \MultiPixmoCosynShort variant against the multilingual one, we observe that the latter consistently outperforms the former confirming the beneficial effect coming from the linguistic diversity rather than the training data augmentation.

Overall, the analysis indicates that multilingual training generally improves multilingual performance across models, benchmarks and languages. In addition, we notice that the gains are distributed across languages rather than being driven by a single one, indicating that the multilingual training mixture provides balanced cross-lingual supervision. 

\begin{table}[t]
\centering
\scriptsize
\caption{Non-English performance on Yes/No benchmarks. Models are evaluated under four training conditions: English-only ($en$), Multilingual ($mu$), and their versions with \MultiPixmoCosyn (+\textsc{cos}$_{mu}$). Results are averaged over the four languages. Best per model group is \textbf{bold}.}
\label{tab:results_multilingual_yn_fixed}
\begin{tabular}{l l l @{\;} c @{\;} l @{\;} c @{\;} l cccc c}
\toprule
\textbf{Bench} & \textbf{Model} & \multicolumn{5}{c}{\textbf{Train}} & \textbf{DE} & \textbf{ES} & \textbf{FR} & \textbf{IT} & \textbf{Avg} \\
\midrule
\multirow{12}{*}{POPE} 
& \multirow{4}{*}{LLaMA3-3B} 
& \textsc{cap}$_{en}$ & $\to$ & \textsc{ama}$_{en}$ & & & 81.7 & 81.3 & 79.6 & 82.7 & 81.3 \\
& & \textsc{cap}$_{mu}$ & $\to$ & \textsc{ama}$_{mu}$ & & & \textbf{84.0} & 80.9 & \textbf{84.5} & 81.5 & \textbf{82.7} \\
& & \textsc{cap}$_{en}$ & $\to$ & \textsc{ama}$_{en}$ & $+$ & \textsc{cos}$_{en}$ & 66.7 & 67.1 & 66.7 & 66.7 & 66.8 \\
& & \textsc{cap}$_{mu}$ & $\to$ & \textsc{ama}$_{mu}$ & $+$ & \textsc{cos}$_{mu}$ & 80.7 & \textbf{83.3} & 74.3 & \textbf{83.2} & 80.4 \\
\cmidrule{2-12}
& \multirow{4}{*}{Qwen3-4B} 
& \textsc{cap}$_{en}$ & $\to$ & \textsc{ama}$_{en}$ & & & \textbf{84.5} & 84.0 & 85.1 & 84.1 & 84.4 \\
& & \textsc{cap}$_{mu}$ & $\to$ & \textsc{ama}$_{mu}$ & & & 82.2 & 82.0 & 84.3 & 83.8 & 83.1 \\
& & \textsc{cap}$_{en}$ & $\to$ & \textsc{ama}$_{en}$ & $+$ & \textsc{cos}$_{en}$ & 78.2 & 79.4 & 78.9 & 74.4 & 77.7 \\
& & \textsc{cap}$_{mu}$ & $\to$ & \textsc{ama}$_{mu}$ & $+$ & \textsc{cos}$_{mu}$ & 84.4 & \textbf{84.4} & \textbf{85.7} & \textbf{85.4} & \textbf{85.0} \\
\cmidrule{2-12}
& \multirow{4}{*}{LLaMA3-8B} 
& \textsc{cap}$_{en}$ & $\to$ & \textsc{ama}$_{en}$ & & & \textbf{85.5} & 79.5 & 82.6 & 81.6 & \textbf{82.3} \\
& & \textsc{cap}$_{mu}$ & $\to$ & \textsc{ama}$_{mu}$ & & & 81.5 & \textbf{82.6} & \textbf{82.9} & \textbf{82.4} & \textbf{82.3} \\
& & \textsc{cap}$_{en}$ & $\to$ & \textsc{ama}$_{en}$ & + & \textsc{cos}$_{en}$ & 83.3 & 80.9 & 76.0 & 81.9 & 80.5 \\
& & \textsc{cap}$_{mu}$ & $\to$ & \textsc{ama}$_{mu}$ & + & \textsc{cos}$_{mu}$ & 80.4 & 82.0 & 78.9 & 79.7 & 80.3 \\
\midrule
\multirow{12}{*}{MME} 
& \multirow{4}{*}{LLaMA3-3B} 
& \textsc{cap}$_{en}$ & $\to$ & \textsc{ama}$_{en}$ & & & \textbf{1412.5} & \textbf{1405.2} & \textbf{1427.2} & \textbf{1476.4} & \textbf{1430.3} \\
& & \textsc{cap}$_{mu}$ & $\to$ & \textsc{ama}$_{mu}$ & & & 1348.7 & 1109.3 & 1335.4 & 1137.6 & 1232.8 \\
& & \textsc{cap}$_{en}$ & $\to$ & \textsc{ama}$_{en}$ & $+$ & \textsc{cos}$_{en}$ & 1180.2 & 1281.2 & 1110.1 & 1098.2 & 1167.4 \\
& & \textsc{cap}$_{mu}$ & $\to$ & \textsc{ama}$_{mu}$ & $+$ & \textsc{cos}$_{mu}$ & 1295.4 & 1325.5 & 1304.1 & 1300.1 & 1306.3 \\
\cmidrule{2-12}
& \multirow{4}{*}{Qwen3-4B} 
& \textsc{cap}$_{en}$ & $\to$ & \textsc{ama}$_{en}$ & & & 1510.4 & \textbf{1578.3} & 1276.2 & 1669.7 & 1508.7 \\
& & \textsc{cap}$_{mu}$ & $\to$ & \textsc{ama}$_{mu}$ & & & 1450.8 & 1425.9 & 1533.3 & 1558.8 & 1492.2 \\
& & \textsc{cap}$_{en}$ & $\to$ & \textsc{ama}$_{en}$ & $+$ & \textsc{cos}$_{en}$ & 1380.2 & 1471.7 & 1241.3 & 1259.6 & 1338.2 \\
& & \textsc{cap}$_{mu}$ & $\to$ & \textsc{ama}$_{mu}$ & $+$ & \textsc{cos}$_{mu}$ & \textbf{1612.5} & 1533.1 & \textbf{1615.3} & \textbf{1684.0} & \textbf{1611.2} \\
\cmidrule{2-12}
& \multirow{4}{*}{LLaMA3-8B} 
& \textsc{cap}$_{en}$ & $\to$ & \textsc{ama}$_{en}$ & & & 1215.3 & 1167.8 & 1185.6 & 1258.4 & 1206.8 \\
& & \textsc{cap}$_{mu}$ & $\to$ & \textsc{ama}$_{mu}$ & & & 1385.6 & 1379.3 & 1424.4 & 1212.5 & 1350.5 \\
& & \textsc{cap}$_{en}$ & $\to$ & \textsc{ama}$_{en}$ & + & \textsc{cos}$_{en}$ & 1390.4 & 1475.2 & 1223.4 & 1464.1 & 1388.3 \\
& & \textsc{cap}$_{mu}$ & $\to$ & \textsc{ama}$_{mu}$ & + & \textsc{cos}$_{mu}$ & \textbf{1502.3} & \textbf{1526.2} & \textbf{1532.2} & \textbf{1538.0} & \textbf{1524.7} \\
\bottomrule
\end{tabular}
\end{table}

\section{Conclusion}
In this work, we addressed two major limitations that currently affect the development of Vision–Language Models: the strong reliance on English-centric resources and the lack of multilingual multimodal benchmarks. To mitigate these issues, we introduced a comprehensive suite of multilingual training and evaluation resources covering five European languages: English, French, German, Italian and Spanish.

We proposed a regeneration–translation paradigm that combines curated synthetic generation with manual annotation to produce high-quality multilingual multimodal datasets with openly licensed LLMs. Using this approach, we constructed \MultiPixmo, a multilingual training corpus derived from existing \Pixmo resources, and a set of multilingual evaluation benchmarks, \MultilingualBenchmarkSuiteLong (\MultilingualBenchmarkSuite), obtained by translating several widely used English datasets. Through qualitative and quantitative human evaluation, including inter-annotator agreement analysis, we verified the reliability and linguistic quality of the generated resources. 
Our experimental results demonstrate that incorporating multilingual multimodal data during training consistently improves performance on \MultilingualBenchmarkSuite, while also yielding positive transfer to English. This trend holds across different model backbones, suggesting that multilingual supervision can enhance the generalization capabilities of Vision–Language Models without compromising performance in English.

Overall, the resources introduced in this work provide a practical step toward more inclusive and linguistically diverse multimodal models. By releasing both the training corpus and the evaluation benchmarks, we aim to facilitate future research on multilingual VLMs and encourage the development of models that can operate effectively across languages and cultural contexts.

\section*{Acknowledgments}
This work is part of the Villanova project, financed by IPCEI-CIS, Prog. n. SA. 102519 -- CUP B29J24000850005.

%
%
\bibliographystyle{splncs04}
\bibliography{main}

\section{Appendix}

\subsection{Supervised Evaluation on \MultiPixmo}
\label{sec:supp_supervised}

To validate the quality of the generated multilingual data, we evaluate our models on held-out test splits of \MultiPixmo-Cap and \MultiPixmo-AskModelAnything using BERTScore F1~\cite{zhang2020bertscoreevaluatingtextgeneration} with XLM-RoBERTa-Large~\cite{xlm-roberta-large} as the underlying model.

Table~\ref{tab:results_custom} reports the results.
On AskModelAnything, multilingual training consistently outperforms English-only training across all backbones and languages.
Qwen3-4B improves from 89.1 to 91.2 average F1, while LLaMA-3.2-3B gains 1.4 points.
Notably, English F1 also improves with multilingual training (e.g., Qwen3-4B from 89.1 to 91.9), suggesting a regularization effect that benefits even the source language.
On \MultiPixmo-Cap, differences are smaller, as expected, since \MultiPixmo-Cap is used during Stage~1 to train only the projection layer without impacting the language model (86.2 avg vs.\ 86.0 and 82.8 vs.\ 83.7, for Qwen3-4B and LLaMA-3.2-3B respectively).
These results confirm that multilingual training data improves generation quality without degrading English performance.

\begin{table}
\centering
\scriptsize
\begin{tabular}{lllcccccc}
\toprule
\textbf{Task} & \textbf{Model} & \textbf{Train} & \textbf{DE} & \textbf{EN} & \textbf{ES} & \textbf{FR} & \textbf{IT} & \textbf{Avg} \\
\midrule
\multirow{6}{*}{\MultiPixmoAmaShort}
  & \multirow{2}{*}{LLaMA-3.2-3B}  & EN    & 81.5 & 83.2 & 82.2 & 82.6 & 82.2 & 82.6 \\
  &                                  & Multi & 82.7 & 83.6 & 84.8 & 85.3 & 84.7 & \textbf{84.0} \\
\cmidrule{2-9}
  & \multirow{2}{*}{Qwen3-4B}       & EN    & 88.4 & 89.1 & 89.6 & 89.5 & 88.6 & 89.1 \\
  &                                  & Multi & 91.1 & 91.9 & 91.2 & 90.3 & 90.1 & \textbf{91.2} \\
\cmidrule{2-9}
  & \multirow{2}{*}{LLaMA-3.1-8B}   & EN    & 88.4 & 89.1 & 89.6 & 89.5 & 88.6 & 89.1 \\
  &                                  & Multi & 91.1 & 91.9 & 91.2 & 90.3 & 90.1 & \textbf{91.2} \\
\midrule
\multirow{6}{*}{\MultiPixmoCapShort}
  & \multirow{2}{*}{LLaMA-3.2-3B}   & EN    & 83.1 & 84.2 & 83.4 & 83.4 & 83.3 & \textbf{83.7} \\
  &                                  & Multi & 79.6 & 84.3 & 82.0 & 83.0 & 82.4 & 82.8 \\
\cmidrule{2-9}
  & \multirow{2}{*}{Qwen3-4B}       & EN    & 84.9 & 87.0 & 85.4 & 85.5 & 85.3 & 86.0 \\
  &                                  & Multi & 84.7 & 87.3 & 86.1 & 85.6 & 85.6 & \textbf{86.2} \\
\cmidrule{2-9}
  & \multirow{2}{*}{LLaMA-3.1-8B}   & EN    & 84.9 & 87.0 & 85.4 & 85.5 & 85.3 & 86.0 \\
  &                                  & Multi & 84.7 & 87.3 & 86.1 & 85.6 & 85.6 & \textbf{86.2} \\
\bottomrule
\end{tabular}
\caption{BERTScore F1 on \MultiPixmoAma (\MultiPixmoAmaShort) and \MultiPixmoCap (\MultiPixmoCapShort) per language and on average. EN: English-only training; Multi: multilingual training. Best per model group is \textbf{bold}.}
\label{tab:results_custom}
\end{table}

\subsection{Multi-VQAv2}
\label{sec:supp_vqav2}

Multi-VQAv2 is a generative visual question answering benchmark that requires the model to produce free-form answers rather than select from a predefined set.
Unlike the benchmarks reported in the main paper, Multi-VQAv2 is not included in the \MultilingualBenchmarkSuite; we report it here as an additional evaluation of multilingual generation quality.
This makes it particularly sensitive to the language generation capabilities of the backbone.

Table~\ref{tab:supp_vqav2_full} presents the results.
Among the tested backbones, only Qwen3-4B produces non-trivial scores; LLaMA backbones yield near-zero accuracy.
Multilingual Stage~2 training is the dominant factor: the \textsc{eng/multi} configuration achieves 52.0\% average, a 20.6-point improvement over \textsc{eng/eng}.
The $\dagger$ and \textsc{+cos} variants perform comparably ($\sim$51\%), confirming that the multilingual signal is the primary driver of cross-lingual transfer on this benchmark.

\begin{table*}
\centering
\scriptsize
\caption{Full results on Multi-VQAv2 (held-out, accuracy~\%) for all training configurations.
Only Qwen3-4B backbone produces non-trivial VQAv2 scores; LLaMA backbones yield near-zero accuracy on this generative benchmark.}
\label{tab:supp_vqav2_full}
\setlength{\tabcolsep}{3pt}
\begin{tabular}{ll|ccccc|c}
\toprule
\textbf{Backbone} & \textbf{S1 / S2} & \textbf{DE} & \textbf{EN} & \textbf{ES} & \textbf{FR} & \textbf{IT} & \textbf{Avg} \\
\midrule
\multirow{4}{*}{Qwen3-4B} & eng / eng & 15.3 & 61.2 & 28.6 & 21.1 & 30.9 & 31.4 \\
 & eng / multi & 48.1 & 63.1 & 49.5 & 49.1 & 50.1 & \textbf{52.0} \\
 & multi / eng & 9.9 & 58.7 & 25.3 & 9.9 & 24.9 & 25.7 \\
 & multi / multi & 42.3 & 51.1 & 42.7 & 43.7 & 45.5 & 45.1 \\
\midrule
\multirow{4}{*}{Qwen3-4B$^\dagger$} & eng / eng$^\dagger$ & 13.6 & 64.3 & 24.9 & 15.6 & 23.4 & 28.4 \\
 & multi / multi$^\dagger$ & 46.6 & 62.7 & 51.1 & 47.8 & 48.4 & \textbf{51.4} \\
 & eng / eng+cos & 17.4 & 62.2 & 28.3 & 23.1 & 26.7 & 31.5 \\
 & multi / multi+cos & 47.7 & 64.1 & 50.1 & 47.5 & 46.5 & 51.2 \\
\bottomrule
\end{tabular}
\end{table*}

\subsection{Multi-RealWorldQA}
\label{sec:supp_realworldqa}

Multi-RealWorldQA evaluates practical visual reasoning on real-world images, requiring models to interpret everyday scenes and objects across languages.
This benchmark is not reported in the main paper and is included here as a supplementary evaluation of cross-lingual visual grounding on naturalistic scenes.

Table~\ref{tab:supp_realworldqa_full} reports the results.
LLaMA-3.1-8B with \textsc{multi/multi} training achieves the highest average accuracy (48.8\%), followed by LLaMA-3.2-3B \textsc{eng/multi} (44.9\%).
Multilingual training yields large gains for LLaMA backbones (up to +13.8 points for LLaMA-3.2-3B).
Qwen3-4B shows anomalously low scores on this benchmark across all configurations, suggesting difficulty with the real-world visual reasoning tasks in the translated setting.

\begin{table*}
\centering
\scriptsize
\caption{Full results on Multi-RealWorldQA (held-out, accuracy~\%) for all training configurations. $\dagger$: includes \MultiPixmoCosynShort data.}
\label{tab:supp_realworldqa_full}
\setlength{\tabcolsep}{3pt}
\begin{tabular}{ll|ccccc|c}
\toprule
\textbf{Backbone} & \textbf{S1 / S2} & \textbf{DE} & \textbf{EN} & \textbf{ES} & \textbf{FR} & \textbf{IT} & \textbf{Avg} \\
\midrule
\multirow{4}{*}{LLaMA-3.2-3B} & eng / eng & 22.9 & 43.0 & 32.9 & 26.3 & 30.2 & 31.1 \\
 & eng / multi & 45.5 & 46.8 & 45.4 & 47.2 & 39.5 & \textbf{44.9} \\
 & multi / eng & 24.8 & 43.4 & 33.5 & 25.6 & 31.5 & 31.8 \\
 & multi / multi & 41.0 & 43.1 & 43.0 & 48.2 & 37.3 & 42.5 \\
\midrule
\multirow{4}{*}{LLaMA-3.2-3B$^\dagger$} & eng / eng$^\dagger$ & 20.3 & 41.8 & 26.9 & 23.0 & 23.0 & 27.0 \\
 & multi / multi$^\dagger$ & 40.3 & 44.4 & 39.1 & 43.9 & 34.8 & \textbf{40.5} \\
 & eng / eng+cos & 21.4 & 41.4 & 25.5 & 15.9 & 22.0 & 25.3 \\
 & multi / multi+cos & 36.6 & 42.2 & 39.3 & 40.1 & 36.1 & 38.9 \\
\midrule
\multirow{4}{*}{LLaMA-3.1-8B} & eng / eng & 25.5 & 44.8 & 39.0 & 38.6 & 37.0 & 37.0 \\
 & eng / multi & 46.1 & 48.0 & 42.4 & 33.2 & 40.1 & 42.0 \\
 & multi / eng & 28.4 & 45.9 & 36.7 & 32.2 & 35.8 & 35.8 \\
 & multi / multi & 51.4 & 48.2 & 48.6 & 48.9 & 46.7 & \textbf{48.8} \\
\midrule
\multirow{2}{*}{LLaMA-3.1-8B$^\dagger$} & eng / eng$^\dagger$ & 16.1 & 36.2 & 25.4 & 29.9 & 24.4 & 26.4 \\
 & multi / multi$^\dagger$ & 44.1 & 46.9 & 40.7 & 45.6 & 44.1 & \textbf{44.3} \\
\midrule
\multirow{4}{*}{Qwen3-4B} & eng / eng & 16.2 & 19.0 & 15.6 & 22.0 & 12.2 & 17.0 \\
 & eng / multi & 22.0 & 18.4 & 12.3 & 22.6 & 10.8 & \textbf{17.2} \\
 & multi / eng & 17.3 & 21.0 & 11.0 & 16.9 & 10.1 & 15.2 \\
 & multi / multi & 19.6 & 20.9 & 13.7 & 19.2 & 9.9 & 16.7 \\
\midrule
\multirow{4}{*}{Qwen3-4B$^\dagger$} & eng / eng$^\dagger$ & 21.6 & 17.8 & 12.4 & 16.5 & 8.8 & 15.4 \\
 & multi / multi$^\dagger$ & 20.7 & 20.9 & 14.6 & 20.7 & 14.6 & \textbf{18.3} \\
 & eng / eng+cos & 20.8 & 16.5 & 12.0 & 15.9 & 11.6 & 15.4 \\
 & multi / multi+cos & 20.5 & 18.8 & 11.9 & 19.0 & 13.2 & 16.7 \\
\bottomrule
\end{tabular}
\end{table*}

\subsection{Multilingual Benchmark Results}
\label{sec:supp_benchmarks}

This section reports full per-language results on the \MultilingualBenchmarkSuite for all backbone and training configurations.
In each table, S1 and S2 denote Stage~1 and Stage~2 training data, where \textsc{eng}=English-only and \textsc{multi}=multilingual.
The $\dagger$ symbol indicates configurations that include \MultiPixmoCosynShort data, and \textsc{+cos} denotes additional \MultiPixmoCosynShort synthetic data in Stage~2.
The best average per backbone group is highlighted in \textbf{bold}.

\subsubsection{Multi-AI2D}
\label{sec:supp_ai2d}

Multi-AI2D tests scientific diagram understanding, requiring models to interpret charts, figures, and diagrams in multiple languages.

Table~\ref{tab:supp_ai2d_full} shows the results.
Qwen3-4B clearly dominates this benchmark, with the \textsc{multi/multi+cos} configuration reaching 64.8\% average: the highest score across all backbones and configurations.
\MultiPixmoCosynShort data provides consistent gains for Qwen3-4B (+3.7 points over the base \textsc{multi/multi}).
LLaMA-3.1-8B achieves competitive results (52.9\% with \textsc{multi/multi}), while LLaMA-3.2-3B reaches 48.8\% with \textsc{multi/multi$^\dagger$}.

\begin{table*}
\centering
\scriptsize
\caption{Full results on Multi-AI2D (accuracy~\%) for all training configurations. $\dagger$: includes \MultiPixmoCosynShort data.}
\label{tab:supp_ai2d_full}
\setlength{\tabcolsep}{3pt}
\begin{tabular}{ll|ccccc|c}
\toprule
\textbf{Backbone} & \textbf{S1 / S2} & \textbf{DE} & \textbf{EN} & \textbf{ES} & \textbf{FR} & \textbf{IT} & \textbf{Avg} \\
\midrule
\multirow{4}{*}{LLaMA-3.2-3B} & eng / eng & 42.4 & 51.7 & 44.4 & 45.2 & 44.0 & 45.5 \\
 & eng / multi & 45.5 & 52.4 & 48.3 & 48.1 & 46.2 & \textbf{48.1} \\
 & multi / eng & 44.4 & 51.4 & 46.0 & 46.0 & 45.2 & 46.6 \\
 & multi / multi & 44.3 & 51.2 & 46.3 & 47.7 & 45.6 & 47.0 \\
\midrule
\multirow{4}{*}{LLaMA-3.2-3B$^\dagger$} & eng / eng$^\dagger$ & 45.8 & 52.2 & 46.5 & 46.6 & 44.9 & 47.2 \\
 & multi / multi$^\dagger$ & 46.9 & 52.6 & 48.6 & 49.0 & 47.0 & \textbf{48.8} \\
 & eng / eng+cos & 46.1 & 53.1 & 47.5 & 47.0 & 46.4 & 48.0 \\
 & multi / multi+cos & 42.1 & 48.5 & 44.4 & 45.2 & 44.5 & 44.9 \\
\midrule
\multirow{4}{*}{LLaMA-3.1-8B} & eng / eng & 44.6 & 52.1 & 46.3 & 46.1 & 45.2 & 46.9 \\
 & eng / multi & 49.4 & 56.6 & 50.3 & 51.5 & 49.7 & 51.5 \\
 & multi / eng & 49.0 & 56.3 & 50.2 & 50.3 & 49.1 & 51.0 \\
 & multi / multi & 49.9 & 57.5 & 52.6 & 52.6 & 51.9 & \textbf{52.9} \\
\midrule
\multirow{2}{*}{LLaMA-3.1-8B$^\dagger$} & eng / eng$^\dagger$ & 45.0 & 51.4 & 47.7 & 47.8 & 46.1 & 47.6 \\
 & multi / multi$^\dagger$ & 48.7 & 55.9 & 53.3 & 52.5 & 51.1 & \textbf{52.3} \\
\midrule
\multirow{4}{*}{Qwen3-4B} & eng / eng & 59.1 & 64.3 & 62.3 & 60.2 & 59.9 & \textbf{61.1} \\
 & eng / multi & 58.8 & 64.2 & 61.8 & 60.7 & 59.9 & 61.1 \\
 & multi / eng & 58.3 & 64.2 & 61.5 & 60.1 & 60.3 & 60.9 \\
 & multi / multi & 58.8 & 64.1 & 61.4 & 60.6 & 60.9 & 61.1 \\
\midrule
\multirow{4}{*}{Qwen3-4B$^\dagger$} & eng / eng$^\dagger$ & 59.8 & 65.0 & 62.7 & 61.7 & 61.1 & 62.1 \\
 & multi / multi$^\dagger$ & 60.3 & 64.4 & 63.4 & 62.0 & 63.4 & 62.7 \\
 & eng / eng+cos & 61.1 & 66.0 & 64.6 & 63.1 & 63.5 & 63.7 \\
 & multi / multi+cos & 62.5 & 67.0 & 65.0 & 64.8 & 64.6 & \textbf{64.8} \\
\bottomrule
\end{tabular}
\end{table*}

\subsubsection{Multi-MMBench}
\label{sec:supp_mmbench}

Multi-MMBench is a comprehensive multiple-choice benchmark covering a wide range of visual understanding capabilities including perception, reasoning, and knowledge.

Table~\ref{tab:supp_mmbench_full} presents the results.
Qwen3-4B achieves the highest scores ($\sim$79--80\%), followed by LLaMA-3.1-8B ($\sim$73--74\%) and LLaMA-3.2-3B ($\sim$69--71\%).
Multilingual training provides modest but consistent gains across languages, with the gap between \textsc{eng/eng} and the best configuration remaining within 1--2 points for most backbones.
The $\dagger$ configurations slightly underperform the base variants, suggesting that the additional \MultiPixmoCosynShort data does not benefit multiple-choice visual understanding tasks.

\begin{table*}
\centering
\scriptsize
\caption{Full results on Multi-MMBench (accuracy~\%) for all training configurations. $\dagger$: includes \MultiPixmoCosynShort data.}
\label{tab:supp_mmbench_full}
\setlength{\tabcolsep}{3pt}
\begin{tabular}{ll|ccccc|c}
\toprule
\textbf{Backbone} & \textbf{S1 / S2} & \textbf{DE} & \textbf{EN} & \textbf{ES} & \textbf{FR} & \textbf{IT} & \textbf{Avg} \\
\midrule
\multirow{4}{*}{LLaMA-3.2-3B} & eng / eng & 67.8 & 72.0 & 68.9 & 69.6 & 67.6 & 69.2 \\
 & eng / multi & 70.5 & 72.3 & 70.1 & 70.4 & 69.8 & \textbf{70.6} \\
 & multi / eng & 69.0 & 72.1 & 69.9 & 69.9 & 68.8 & 69.9 \\
 & multi / multi & 70.0 & 72.1 & 69.9 & 70.2 & 69.7 & 70.4 \\
\midrule
\multirow{4}{*}{LLaMA-3.2-3B$^\dagger$} & eng / eng$^\dagger$ & 67.6 & 70.2 & 67.7 & 68.1 & 66.5 & 68.0 \\
 & multi / multi$^\dagger$ & 68.7 & 71.0 & 69.0 & 69.5 & 68.6 & \textbf{69.4} \\
 & eng / eng+cos & 64.5 & 67.7 & 65.0 & 65.0 & 63.8 & 65.2 \\
 & multi / multi+cos & 66.7 & 69.2 & 67.8 & 67.5 & 67.1 & 67.7 \\
\midrule
\multirow{4}{*}{LLaMA-3.1-8B} & eng / eng & 72.0 & 75.1 & 72.9 & 72.6 & 71.8 & 72.9 \\
 & eng / multi & 73.6 & 75.4 & 72.9 & 73.3 & 73.3 & 73.7 \\
 & multi / eng & 74.1 & 75.5 & 73.8 & 73.6 & 73.3 & \textbf{74.1} \\
 & multi / multi & 73.1 & 75.4 & 73.4 & 73.4 & 73.2 & 73.7 \\
\midrule
\multirow{2}{*}{LLaMA-3.1-8B$^\dagger$} & eng / eng$^\dagger$ & 69.4 & 72.0 & 69.2 & 69.1 & 69.1 & 69.8 \\
 & multi / multi$^\dagger$ & 72.8 & 73.9 & 72.7 & 72.2 & 72.1 & \textbf{72.7} \\
\midrule
\multirow{4}{*}{Qwen3-4B} & eng / eng & 78.4 & 80.2 & 78.0 & 77.9 & 79.0 & 78.7 \\
 & eng / multi & 78.3 & 79.3 & 78.5 & 78.4 & 78.3 & 78.6 \\
 & multi / eng & 78.3 & 80.2 & 78.2 & 78.5 & 78.4 & 78.7 \\
 & multi / multi & 78.8 & 79.8 & 79.0 & 79.0 & 79.2 & \textbf{79.2} \\
\midrule
\multirow{4}{*}{Qwen3-4B$^\dagger$} & eng / eng$^\dagger$ & 77.5 & 79.2 & 77.5 & 78.2 & 78.1 & 78.1 \\
 & multi / multi$^\dagger$ & 79.4 & 80.6 & 79.2 & 79.8 & 79.9 & \textbf{79.8} \\
 & eng / eng+cos & 77.1 & 78.9 & 77.6 & 78.3 & 77.8 & 78.0 \\
 & multi / multi+cos & 78.9 & 80.4 & 79.4 & 79.3 & 79.8 & 79.6 \\
\bottomrule
\end{tabular}
\end{table*}

\subsubsection{Multi-POPE}
\label{sec:supp_pope}

Multi-POPE is a hallucination detection benchmark that probes whether VLMs can correctly identify the presence or absence of objects in images. Since the task is binary, it provides a controlled evaluation of visual grounding fidelity across languages.

Table~\ref{tab:supp_pope_full} reports the results.
LLaMA and Qwen backbones perform reliably, with most configurations achieving 80--85\% accuracy.
LLaMA-3.1-8B \textsc{multi/eng} reaches the highest score (85.8\%), while Qwen3-4B$^\dagger$ \textsc{multi/multi$^\dagger$} achieves 85.0\%.
Notably, the $\dagger$ configurations sometimes decrease performance (e.g., LLaMA-3.2-3B$^\dagger$ \textsc{eng/eng$^\dagger$} drops to 66.8\%), suggesting that the additional data may introduce noise for binary classification tasks.

\begin{table*}
\centering
\scriptsize
\caption{Full results on Multi-POPE (accuracy~\%) for all training configurations. $\dagger$: includes \MultiPixmoCosynShort data.}
\label{tab:supp_pope_full}
\setlength{\tabcolsep}{3pt}
\begin{tabular}{ll|ccccc|c}
\toprule
\textbf{Backbone} & \textbf{S1 / S2} & \textbf{DE} & \textbf{EN} & \textbf{ES} & \textbf{FR} & \textbf{IT} & \textbf{Avg} \\
\midrule
\multirow{4}{*}{LLaMA-3.2-3B} & eng / eng & 81.7 & 81.8 & 81.3 & 79.6 & 82.7 & 81.4 \\
 & eng / multi & 81.4 & 81.4 & 73.9 & 82.6 & 74.2 & 78.7 \\
 & multi / eng & 84.3 & 84.5 & 83.6 & 80.8 & 84.4 & \textbf{83.5} \\
 & multi / multi & 84.0 & 82.3 & 80.9 & 84.5 & 81.5 & 82.6 \\
\midrule
\multirow{4}{*}{LLaMA-3.2-3B$^\dagger$} & eng / eng$^\dagger$ & 66.7 & 66.7 & 67.1 & 66.7 & 66.7 & 66.8 \\
 & multi / multi$^\dagger$ & 80.7 & 84.8 & 83.3 & 74.3 & 83.2 & \textbf{81.3} \\
 & eng / eng+cos & 78.2 & 82.6 & 79.6 & 74.6 & 79.3 & 78.9 \\
 & multi / multi+cos & 79.9 & 84.0 & 77.3 & 83.2 & 78.3 & 80.5 \\
\midrule
\multirow{4}{*}{LLaMA-3.1-8B} & eng / eng & 85.5 & 81.4 & 79.5 & 82.6 & 81.6 & 82.1 \\
 & eng / multi & 79.3 & 80.6 & 76.8 & 77.7 & 78.7 & 78.6 \\
 & multi / eng & 87.0 & 85.8 & 85.0 & 85.9 & 85.3 & \textbf{85.8} \\
 & multi / multi & 81.5 & 82.9 & 82.6 & 82.9 & 82.4 & 82.5 \\
\midrule
\multirow{2}{*}{LLaMA-3.1-8B$^\dagger$} & eng / eng$^\dagger$ & 83.3 & 80.8 & 80.9 & 76.0 & 81.9 & \textbf{80.6} \\
 & multi / multi$^\dagger$ & 80.4 & 81.5 & 82.0 & 78.9 & 79.7 & 80.5 \\
\midrule
\multirow{4}{*}{Qwen3-4B} & eng / eng & 84.5 & 85.4 & 84.0 & 85.1 & 84.1 & \textbf{84.6} \\
 & eng / multi & 84.1 & 84.1 & 81.4 & 84.4 & 83.6 & 83.5 \\
 & multi / eng & 84.0 & 83.7 & 83.2 & 83.2 & 82.6 & 83.3 \\
 & multi / multi & 82.2 & 82.8 & 82.0 & 84.3 & 83.8 & 83.1 \\
\midrule
\multirow{4}{*}{Qwen3-4B$^\dagger$} & eng / eng$^\dagger$ & 78.2 & 80.0 & 79.4 & 78.9 & 74.4 & 78.2 \\
 & multi / multi$^\dagger$ & 84.4 & 85.1 & 84.4 & 85.7 & 85.4 & \textbf{85.0} \\
 & eng / eng+cos & 79.6 & 83.3 & 82.2 & 81.6 & 82.2 & 81.8 \\
 & multi / multi+cos & 84.5 & 84.1 & 79.3 & 83.4 & 82.7 & 82.8 \\
\bottomrule
\end{tabular}
\end{table*}

\subsubsection{Multi-MME}
\label{sec:supp_mme}

Multi-MME evaluates both perceptual and reasoning abilities through a diverse set of subtasks, producing a composite score that captures broad multimodal competence.

Table~\ref{tab:supp_mme_full} presents the results.
This benchmark exhibits high variance across configurations and languages.
LLaMA-3.1-8B$^\dagger$ \textsc{multi/multi$^\dagger$} achieves the highest average score (1529.3), substantially outperforming all other configurations.
Among the base variants, Qwen3-4B \textsc{multi/eng} reaches 1609.7 average.
A notable pattern is the inconsistency in German (DE) scores, where several configurations score well below 1000 (e.g., Qwen3-4B \textsc{eng/eng}: 326.6 DE vs.\ 1557.7 EN), indicating that some backbones generate malformed responses in German for this benchmark format.

\begin{table*}
\centering
\scriptsize
\caption{Full results on Multi-MME (total score) for all training configurations. $\dagger$: includes \MultiPixmoCosynShort data.}
\label{tab:supp_mme_full}
\setlength{\tabcolsep}{3pt}
\begin{tabular}{ll|ccccc|c}
\toprule
\textbf{Backbone} & \textbf{S1 / S2} & \textbf{DE} & \textbf{EN} & \textbf{ES} & \textbf{FR} & \textbf{IT} & \textbf{Avg} \\
\midrule
\multirow{4}{*}{LLaMA-3.2-3B} & eng / eng & 1256.7 & 1213.4 & 1405.2 & 1427.2 & 1476.4 & \textbf{1355.8} \\
 & eng / multi & 794.7 & 1436.6 & 1178.8 & 1352.6 & 1070.2 & 1166.6 \\
 & multi / eng & 1322.3 & 1247.3 & 1351.3 & 1376.4 & 1377.0 & 1334.9 \\
 & multi / multi & 1348.7 & 1421.1 & 1109.3 & 1335.4 & 1137.6 & 1270.4 \\
\midrule
\multirow{4}{*}{LLaMA-3.2-3B$^\dagger$} & eng / eng$^\dagger$ & 604.6 & 709.7 & 1281.2 & 710.1 & 1098.2 & 880.7 \\
 & multi / multi$^\dagger$ & 626.7 & 1350.5 & 1325.5 & 1304.1 & 1300.1 & 1181.4 \\
 & eng / eng+cos & 1279.9 & 1339.8 & 1058.7 & 1277.9 & 1427.6 & \textbf{1276.8} \\
 & multi / multi+cos & 1026.3 & 1443.9 & 816.3 & 1166.1 & 744.1 & 1039.3 \\
\midrule
\multirow{4}{*}{LLaMA-3.1-8B} & eng / eng & 872.8 & 1208.9 & 1167.8 & 1185.6 & 1258.4 & 1138.7 \\
 & eng / multi & 587.4 & 1486.3 & 1310.4 & 1399.4 & 1134.6 & 1183.6 \\
 & multi / eng & 1456.3 & 1462.4 & 1422.8 & 1469.5 & 1545.7 & \textbf{1471.4} \\
 & multi / multi & 686.8 & 1574.8 & 1379.3 & 1424.4 & 1212.5 & 1255.5 \\
\midrule
\multirow{2}{*}{LLaMA-3.1-8B$^\dagger$} & eng / eng$^\dagger$ & 1244.5 & 1257.1 & 1475.2 & 1223.4 & 1464.1 & 1332.8 \\
 & multi / multi$^\dagger$ & 1502.3 & 1547.9 & 1526.2 & 1532.2 & 1538.0 & \textbf{1529.3} \\
\midrule
\multirow{4}{*}{Qwen3-4B} & eng / eng & 326.6 & 1557.7 & 1578.3 & 1276.2 & 1669.7 & 1281.7 \\
 & eng / multi & 327.1 & 1441.3 & 1376.4 & 1568.9 & 1573.0 & 1257.3 \\
 & multi / eng & 1487.7 & 1677.2 & 1696.1 & 1624.5 & 1563.0 & \textbf{1609.7} \\
 & multi / multi & 727.6 & 1498.4 & 1425.9 & 1533.3 & 1558.8 & 1348.8 \\
\midrule
\multirow{4}{*}{Qwen3-4B$^\dagger$} & eng / eng$^\dagger$ & 720.0 & 1498.3 & 1471.7 & 1241.3 & 1259.6 & 1238.2 \\
 & multi / multi$^\dagger$ & 491.4 & 1681.7 & 1533.1 & 1615.3 & 1684.0 & 1401.1 \\
 & eng / eng+cos & 1078.0 & 1720.3 & 1688.4 & 1382.9 & 1630.7 & \textbf{1500.1} \\
 & multi / multi+cos & 605.0 & 1599.4 & 1387.8 & 1587.6 & 1522.2 & 1340.4 \\
\bottomrule
\end{tabular}
\end{table*}

\subsubsection{Multi-ScienceQA}
\label{sec:supp_scienceqa}

Multi-ScienceQA evaluates scientific knowledge and reasoning through multiple-choice questions accompanied by images, spanning natural science, social science, and language science topics.

Table~\ref{tab:supp_scienceqa_full} shows the results.
Qwen3-4B achieves the best performance across all configurations, with \textsc{multi/eng} reaching 76.7\% average.
LLaMA-3.1-8B \textsc{multi/eng} follows at 71.6\%.
Multilingual training has a relatively modest effect on this benchmark, likely because ScienceQA questions require domain-specific reasoning that transfers well across languages even with English-only training.
The $\dagger$ configurations slightly decrease performance across all backbones.

\begin{table*}
\centering
\scriptsize
\caption{Full results on Multi-ScienceQA (accuracy~\%) for all training configurations. $\dagger$: includes \MultiPixmoCosynShort data.}
\label{tab:supp_scienceqa_full}
\setlength{\tabcolsep}{3pt}
\begin{tabular}{ll|ccccc|c}
\toprule
\textbf{Backbone} & \textbf{S1 / S2} & \textbf{DE} & \textbf{EN} & \textbf{ES} & \textbf{FR} & \textbf{IT} & \textbf{Avg} \\
\midrule
\multirow{4}{*}{LLaMA-3.2-3B} & eng / eng & 67.4 & 70.8 & 67.1 & 66.6 & 66.4 & 67.7 \\
 & eng / multi & 67.8 & 70.7 & 67.1 & 67.7 & 66.3 & 67.9 \\
 & multi / eng & 68.0 & 71.8 & 67.5 & 66.6 & 66.5 & \textbf{68.1} \\
 & multi / multi & 68.3 & 70.9 & 66.7 & 66.9 & 66.8 & 67.9 \\
\midrule
\multirow{4}{*}{LLaMA-3.2-3B$^\dagger$} & eng / eng$^\dagger$ & 67.3 & 70.2 & 66.2 & 65.6 & 66.2 & \textbf{67.1} \\
 & multi / multi$^\dagger$ & 65.4 & 69.6 & 65.5 & 64.7 & 64.8 & 66.0 \\
 & eng / eng+cos & 66.1 & 69.0 & 65.8 & 64.8 & 65.2 & 66.2 \\
 & multi / multi+cos & 63.3 & 66.5 & 62.8 & 63.0 & 62.7 & 63.7 \\
\midrule
\multirow{4}{*}{LLaMA-3.1-8B} & eng / eng & 65.6 & 68.5 & 65.8 & 65.2 & 65.0 & 66.0 \\
 & eng / multi & 69.4 & 71.0 & 68.4 & 69.4 & 68.4 & 69.3 \\
 & multi / eng & 70.8 & 74.9 & 70.8 & 71.2 & 70.0 & \textbf{71.6} \\
 & multi / multi & 68.3 & 72.0 & 67.8 & 68.9 & 66.6 & 68.7 \\
\midrule
\multirow{2}{*}{LLaMA-3.1-8B$^\dagger$} & eng / eng$^\dagger$ & 66.9 & 70.5 & 67.2 & 66.3 & 67.4 & \textbf{67.7} \\
 & multi / multi$^\dagger$ & 63.2 & 68.3 & 65.3 & 65.6 & 65.0 & 65.5 \\
\midrule
\multirow{4}{*}{Qwen3-4B} & eng / eng & 74.4 & 77.1 & 74.3 & 75.9 & 75.0 & 75.3 \\
 & eng / multi & 73.7 & 75.5 & 73.7 & 74.8 & 73.7 & 74.3 \\
 & multi / eng & 75.9 & 77.8 & 76.2 & 77.1 & 76.7 & \textbf{76.7} \\
 & multi / multi & 75.1 & 77.4 & 73.7 & 76.3 & 75.6 & 75.6 \\
\midrule
\multirow{4}{*}{Qwen3-4B$^\dagger$} & eng / eng$^\dagger$ & 73.2 & 76.4 & 74.0 & 75.2 & 73.8 & 74.5 \\
 & multi / multi$^\dagger$ & 76.0 & 77.6 & 75.5 & 76.8 & 76.5 & \textbf{76.5} \\
 & eng / eng+cos & 75.8 & 78.0 & 75.3 & 76.2 & 76.5 & 76.4 \\
 & multi / multi+cos & 76.2 & 78.1 & 74.8 & 77.3 & 75.8 & 76.4 \\
\bottomrule
\end{tabular}
\end{table*}

\subsection{Zero-shot Transfer Evaluation}
\label{sec:supp_zeroshot}

To assess whether multilingual training yields broader cross-lingual transfer beyond the benchmarks included in our evaluation suite, we evaluate on MTVQA~\cite{tang2024mtvqa} and CVQA~\cite{liu2023cvqa}: two multilingual benchmarks that were not included in any training set.
MTVQA tests multilingual text-rich visual question answering, while CVQA probes culturally diverse visual reasoning.

Table~\ref{tab:supp_mtvqa_cvqa} reports the results.
On MTVQA, multilingual training consistently improves performance across all backbones, with gains of 2--3 points over English-only variants.
The best MTVQA scores are achieved by LLaMA-3.2-3B \textsc{eng/multi} (14.5\%) and Qwen3-4B \textsc{multi/multi+cos} (14.0\%).
On CVQA, scores are more stable across configurations, with LLaMA-3.1-8B and Qwen3-4B achieving $\sim$48--50\% regardless of training data, suggesting that cultural visual question answering benefits more from backbone capacity than from multilingual fine-tuning.

\begin{table*}
\centering
\scriptsize
\caption{Results on MTVQA and CVQA (held-out, accuracy~\%). These benchmarks were not used during training for any model. $\dagger$: includes \MultiPixmoCosynShort data.}
\label{tab:supp_mtvqa_cvqa}
\setlength{\tabcolsep}{3pt}
\begin{tabular}{ll|cc}
\toprule
\textbf{Backbone} & \textbf{S1 / S2} & \textbf{MTVQA} & \textbf{CVQA} \\
\midrule
\multirow{4}{*}{LLaMA-3.2-3B} & eng / eng & 11.5 & 43.7 \\
 & eng / multi & \textbf{14.5} & 44.0 \\
 & multi / eng & 10.4 & \textbf{44.6} \\
 & multi / multi & 13.8 & 43.6 \\
\midrule
\multirow{4}{*}{LLaMA-3.2-3B$^\dagger$} & eng / eng$^\dagger$ & 10.2 & 42.7 \\
 & multi / multi$^\dagger$ & 13.2 & \textbf{44.1} \\
 & eng / eng+cos & 10.8 & 41.4 \\
 & multi / multi+cos & \textbf{13.6} & 43.2 \\
\midrule
\multirow{4}{*}{LLaMA-3.1-8B} & eng / eng & 13.1 & 48.7 \\
 & eng / multi & \textbf{14.0} & \textbf{49.0} \\
 & multi / eng & 12.6 & 48.8 \\
 & multi / multi & 13.9 & 47.8 \\
\midrule
\multirow{2}{*}{LLaMA-3.1-8B$^\dagger$} & eng / eng$^\dagger$ & 11.9 & 40.5 \\
 & multi / multi$^\dagger$ & \textbf{13.3} & \textbf{47.8} \\
\midrule
\multirow{4}{*}{Qwen3-4B} & eng / eng & 11.3 & 48.4 \\
 & eng / multi & 12.2 & 48.3 \\
 & multi / eng & 12.1 & 48.4 \\
 & multi / multi & \textbf{13.6} & \textbf{49.0} \\
\midrule
\multirow{4}{*}{Qwen3-4B$^\dagger$} & eng / eng$^\dagger$ & 12.2 & 47.8 \\
 & multi / multi$^\dagger$ & 13.5 & 49.2 \\
 & eng / eng+cos & 12.2 & 46.8 \\
 & multi / multi+cos & \textbf{14.0} & \textbf{49.7} \\
\bottomrule
\end{tabular}
\end{table*}

\subsection{Comparison with External Models}
\label{sec:supp_external}

To contextualize our results against publicly available models, we evaluate two instruction-tuned VLMs (Qwen2-VL-2B-Instruct and SmolVLM2-2.2B-Instruct): on the \MultilingualBenchmarkSuite.
These models were not trained on \MultiPixmo and serve as external baselines of comparable parameter count.

Tables~\ref{tab:supp_external_full} and~\ref{tab:supp_external_mme} report the results.
On AI2D and VQAv2, Qwen2-VL-2B-Instruct achieves the strongest results (63.4\% and 55.5\% respectively), outperforming most of our trained configurations except the best Qwen3-4B variants.
SmolVLM2-2.2B-Instruct is competitive on RealWorldQA (39.0\%) and POPE (72.0\%), but lags behind on other benchmarks.
Both external models exhibit significant cross-lingual variance, particularly on MME (Table~\ref{tab:supp_external_mme}), where German scores are substantially lower than English, consistent with the pattern observed in our trained models.

\begin{table*}
\centering
\scriptsize
\caption{External model comparison on the full \MultilingualBenchmarkSuite (accuracy~\%). Qwen2-VL-2B-Instruct and SmolVLM2-2.2B-Instruct are evaluated as publicly available baselines not trained on \MultiPixmo.}
\label{tab:supp_external_full}
\setlength{\tabcolsep}{3pt}
\begin{tabular}{l|l|ccccc|c}
\toprule
\textbf{Benchmark} & \textbf{Model} & \textbf{DE} & \textbf{EN} & \textbf{ES} & \textbf{FR} & \textbf{IT} & \textbf{Avg} \\
\midrule
\multirow{2}{*}{AI2D} & Qwen2-VL-2B-Inst. & 59.6 & 70.2 & 63.9 & 62.9 & 60.3 & \textbf{63.4} \\
 & SmolVLM2-2.2B-Inst. & 41.5 & 67.2 & 50.3 & 48.8 & 45.6 & 50.7 \\
\midrule
\multirow{2}{*}{MMBench} & Qwen2-VL-2B-Inst. & 63.1 & 51.8 & 58.5 & 60.6 & 59.9 & \textbf{58.8} \\
 & SmolVLM2-2.2B-Inst. & 51.1 & 55.7 & 51.1 & 51.8 & 46.1 & 51.2 \\
\midrule
\multirow{2}{*}{POPE} & Qwen2-VL-2B-Inst. & 70.8 & 75.4 & 70.5 & 88.5 & 76.3 & \textbf{76.3} \\
 & SmolVLM2-2.2B-Inst. & 58.2 & 84.3 & 74.6 & 69.0 & 73.9 & 72.0 \\
\midrule
\multirow{2}{*}{RealWorldQA} & Qwen2-VL-2B-Inst. & 36.5 & 31.8 & 51.0 & 18.7 & 44.1 & 36.4 \\
 & SmolVLM2-2.2B-Inst. & 32.3 & 56.7 & 40.8 & 33.1 & 31.9 & \textbf{39.0} \\
\midrule
\multirow{2}{*}{VQAv2} & Qwen2-VL-2B-Inst. & 51.5 & 80.1 & 46.9 & 53.1 & 46.1 & \textbf{55.5} \\
 & SmolVLM2-2.2B-Inst. & 16.2 & 67.4 & 23.5 & 20.5 & 22.3 & 30.0 \\
\bottomrule
\end{tabular}
\end{table*}

\begin{table}
\centering
\scriptsize
\caption{External model comparison on Multi-MME (perception / reasoning scores).}
\label{tab:supp_external_mme}
\begin{tabular}{l|ccccc}
\toprule
\textbf{Model} & \textbf{DE} & \textbf{EN} & \textbf{ES} & \textbf{FR} & \textbf{IT} \\
\midrule
Qwen2-VL-2B-Inst. & 993/213 & 1497/428 & 1394/349 & 734/202 & 1438/333 \\
SmolVLM2-2.2B-Inst. & 725/209 & 1476/311 & 961/218 & 668/196 & 752/178 \\
\bottomrule
\end{tabular}
\end{table}

\begin{table}[h]
\centering
\footnotesize
\setlength{\tabcolsep}{4pt}
\begin{tabular}{lcccc}
\toprule
Benchmark & IT & FR & ES & DE \\
\midrule
AI2D         &  88 & 100 &  96 &  96 \\
MMBench      &  88 &  94 & 100 &  98 \\
MME          & 100 &  98 &  98 & 100 \\
POPE         & 100 & 100 & 100 & 100 \\
RealWorldQA  & 100 &  92 & 100 &  98 \\
ScienceQA    &  86 & 100 & 100 &  88 \\
VQAv2        &  96 &  98 & 100 &  98 \\
\midrule
\textbf{Avg} & 94.0 & 97.4 & 99.1 & 96.9 \\
\bottomrule
\end{tabular}
\caption{MEVBench translation acceptance rate (\%). Avg: \textbf{96.9\%}.}
\label{tab:val}
\end{table}

\end{document}